\documentclass[sigconf]{acmart}

\AtBeginDocument{%
  }

\copyrightyear{2026}
\acmYear{2026}
\setcopyright{cc}
\setcctype{by}
\acmConference[KDD '26]{Proceedings of the 32nd ACM SIGKDD Conference on Knowledge Discovery and Data Mining V.2}{August 09--13, 2026}{Jeju Island, Republic of Korea}
\acmBooktitle{Proceedings of the 32nd ACM SIGKDD Conference on Knowledge Discovery and Data Mining V.2 (KDD '26), August 09--13, 2026, Jeju Island, Republic of Korea}
\acmDOI{10.1145/3770855.3817814}
\acmISBN{979-8-4007-2259-2/2026/08}

\usepackage{xspace}
\newcommand{\model}{GRiD\xspace}
\usepackage{multirow}
\usepackage{graphicx}    
\usepackage{subcaption}  
\usepackage{algorithm}
\usepackage{algorithmic}
\usepackage{xurl}
\begin{document}

\title{Generating Graph-Like Logical Rules for Knowledge Graph Reasoning via Diffusion Models}

\author{Haoxiang Cheng}
\authornote{These authors contributed equally to this research.}
\affiliation{%
  \institution{Laboratory for Big Data and Decision, National University of Defense Technology}
  \city{Changsha}
  \country{China}
}
\email{hx_chenggfkd@nudt.edu.cn}

\author{Yunfei Wang}
\authornotemark[1]
\affiliation{%
  \institution{National Key Laboratory of Information Systems Engineering, National University of Defense Technology}
  \city{Changsha}
  \country{China}
}
\email{wangyunfei@nudt.edu.cn}

\author{Chao Chen}
\authornotemark[1]
\affiliation{%
  \institution{Laboratory for Big Data and Decision, National University of Defense Technology}
  \city{Changsha}
  \country{China}
}
\email{chenc1997@nudt.edu.cn}

\author{Kewei Cheng}
\authornote{Project Leader.}
\affiliation{%
\institution{Microsoft Corporation}
  \city{Redmond}
\country{USA}
}
\email{keweicheng@microsoft.com}

\author{Zhipeng Lin}
\affiliation{%
\institution{College of Computer Science and Technology, National University of Defense Technology}
\city{Changsha}
\country{China}
}
\email{linzhipeng13@nudt.edu.cn}

\author{Haoxuan Li}
\affiliation{%
  \institution{Peking University}
  \city{Beijing}
  \country{China}
}
\email{hxli@stu.pku.edu.cn}

\author{Changjun Fan}
\authornote{Corresponding authors.}
\affiliation{%
  \institution{Laboratory for Big Data and Decision, National University of Defense Technology}
  \city{Changsha}
  \country{China}
}
\email{fanchangjun@nudt.edu.cn}

\author{Shixuan Liu}
\authornotemark[3]
\affiliation{%
\institution{College of Computer Science and Technology, National University of Defense Technology}
\city{Changsha}
\country{China}
}
\email{szftandy@hotmail.com}

\renewcommand{\shortauthors}{Haoxiang Cheng et al.}

\begin{abstract}

Logical rules constitute a cornerstone of knowledge graph (KG) reasoning, valued for their interpretability and ability to model relational patterns. However, existing rule mining methods predominantly focus on simple chain-like rules and therefore neglect the richer relational information encoded in graph-like structures, such as cycles and branches. This limitation is further exacerbated by computational bottlenecks caused by the combinatorial explosion of the search space, which is especially challenging for graph-like rules. Meanwhile, generative approaches such as diffusion models, despite their success in other domains, cannot be directly applied to rule mining because their training objectives are not aligned with the goal of learning high-quality rules, and non-differentiable KG rule quality metrics cannot directly guide model optimization. To address these limitations, we propose GRiD, a framework that reformulates graph-like rule discovery as a discrete generative process conditioned on the target relation. GRiD employs a two-phase training strategy. First, supervised pre-training enables GRiD to capture structural priors from subgraphs sampled from the KG meta-graph. Subsequently, reinforcement learning is applied to fine-tune GRiD through policy gradient optimization guided directly by non-differentiable rule-quality metrics. Experiments on six benchmark datasets show that GRiD achieves competitive performance on KG completion tasks. Ablation studies confirm the efficiency and robustness of GRiD and further show that graph-like rules complement chain-like rules in KG completion. Our code and datasets are available in \url{https://github.com/Haoxiang-Cheng/GRiD}.
\end{abstract}





\maketitle

\section{Introduction}

\begin{figure*}[t]
\centering
\includegraphics[width=\linewidth]{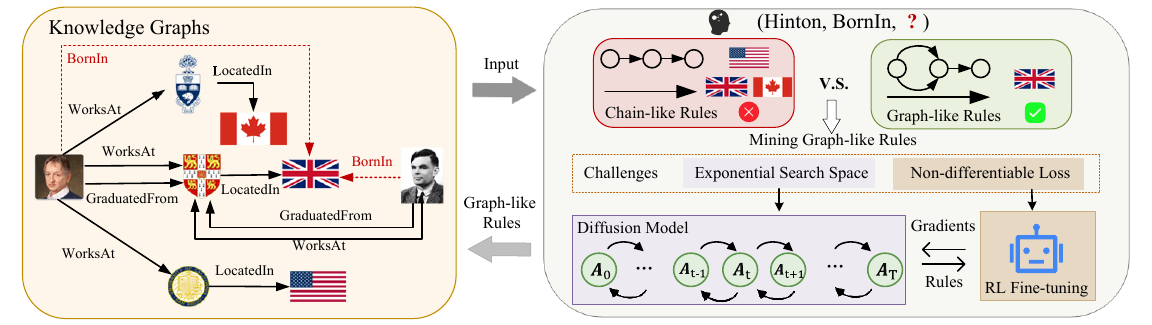}
\caption{While existing rule-based methods predominantly rely on chain-like structures, such structures are insufficient for capturing complex interconnected patterns in KGs. Graph-like rules provide a more expressive alternative but are hindered by the exponential growth of the search space. Therefore, we propose a diffusion-based method for efficient rule generation. However, the standard supervised reconstruction objective for these models is misaligned with the goal of discovering semantically high-quality rules. We therefore introduce RL fine-tuning to directly optimize the generation policy using rewards based on structural quality, enabling the efficient generation of high-quality graph-like rules.}
\label{fig1_introduction}
\end{figure*}

Knowledge graphs (KGs) represent knowledge as factual triples $(e_h, r, e_t)$ and serve as critical components in intelligent systems such as semantic search and question answering~\citep{yin2025efokcqa, saxena-etal-2022-sequence, sun2025kerag, zhang2024question}. However, KGs are inherently incomplete, necessitating reasoning methods to infer missing facts. Among various paradigms, rule-based reasoning offers interpretability by providing explicit logical rules that capture relational dependencies and enable transparent inference~\citep{ji2021survey}. A logical rule typically expresses an implication of the form $\rho:\rho_h \leftarrow \rho_b$, where the rule head $\rho_h$ can be inferred from a conjunction of body atoms $\rho_b$. In practice, rules are often induced from observed path instances, creating a direct link between schema semantics and instance-level evidence. For example, as illustrated in Fig.~\ref{fig1_introduction}, the rule $\rho_1: \textit{BornIn}(x,z) \leftarrow \textit{WorksAt}(x,y) \wedge \textit{LocatedIn}(y,z)$ is induced from the instance path \textit{BornIn(Turing, UK)} $\leftarrow$ \textit{WorksAt(Turing, Cambridge)} $\wedge$ \textit{LocatedIn(Cambridge, UK)}. Such instance-grounded induction yields transparent and verifiable reasoning paths, highlighting the utility of rule-based approaches.

Despite their transparency, traditional rule-based methods are largely confined to chain-like rules, which are sequences of relations connected by shared variables~\citep{AMIE2013luis,cheng2023NCRL,AnyBURL2023}. While chain-like rules are efficient to mine and apply, their linear structure limits their ability to express richer relational patterns, such as conjunctive or overlapping constraints. This restriction often leads to ambiguous predictions in KGs with complex schemas. For instance, applying the chain-like rule $\rho_1$ to infer \textit{BornIn}(Hinton, $?$) may yield multiple candidates, because the rule cannot distinguish among multiple candidate workplaces associated with the subject. This limitation arises from the inability of chain-like rules to jointly enforce multi-relational constraints, highlighting the need for more discriminative rule structures.

Graph-like rules, whose bodies form directed graphs rather than simple chains, can address this limitation by enforcing multiple relational constraints simultaneously. For instance, a graph-like rule such as $\rho_2:\textit{BornIn}(x, z) \leftarrow \textit{WorksAt}(x, y) \wedge \textit{GraduatedFrom}(x, y) \wedge \textit{LocatedIn}(y, z)$ incorporates joint evidence through shared variables, thereby reducing ambiguity. However, extending rule learning from chain-like rules to general graph-like structures introduces a fundamental computational bottleneck. Even mining simple chain-like rules involves a combinatorial exponential search space, typically on the order of $O(|R|^L)$ for length $L$. Graph-like rules, which generalize chains by introducing non-linear dependencies, further expand this space to $O((L|R|)^{L})$. This issue is inherent in search-based paradigms, particularly those extracting rules from instance paths, where traversing large KG instances incurs substantial computational overhead. Consequently, this challenge motivates a shift from iterative search to more efficient exploration of the rule space, in line with recent studies that formulate graph-structured combinatorial optimization as learnable optimization procedures~\citep{pu2024solving}.

While search-based rule discovery is constrained by combinatorial complexity, generative approaches have recently emerged as an alternative. Diffusion models, in particular, have shown effectiveness in generating structured data while maintaining robustness and diversity. Unlike Generative Adversarial Networks and Variational Autoencoders, which may suffer from mode collapse or over-smoothing~\citep{barannikov2021manifold}, diffusion models can provide more stable coverage of complex output spaces. Moreover, while LLMs remain vulnerable to logical inconsistency, hallucination, and symbolic-translation errors in complex reasoning tasks~\cite{cheng2025empowering,chen2026logiconbench,yang2026mad, EvoPath2025}, diffusion models can incorporate explicit structural constraints to preserve rule validity. Nevertheless, existing diffusion-based methods in the KG domain primarily focus on instance-level tasks such as fact completion rather than on discovering explicit schema-level logical patterns~\citep{long2024fact, long2024kgdm,zhou2024unifying}.

Adapting diffusion models from instance-level reconstruction to schema-level rule induction is nontrivial. A central challenge lies in the non-differentiable nature of rule quality metrics. Unlike standard diffusion training, which optimizes differentiable reconstruction losses, logical rules are typically assessed using discrete quality metrics such as confidence and coverage. These metrics cannot be directly back-propagated, creating a mismatch between the training objective and the semantic quality of generated rules. As a result, naively applying diffusion models may fail to align generation with semantically meaningful rules.

To address these challenges, we propose~\model, a framework for generating \textbf{\underline{G}}raph-like \textbf{\underline{R}}ules w\textbf{\underline{i}}th \textbf{\underline{D}}iffusion models for KG reasoning.~\model~reformulates rule discovery as a conditional discrete diffusion process over the rule-body adjacency matrix. To overcome the non-differentiability of discrete rule metrics,~\model~incorporates Reinforcement Learning (RL) to fine-tune the diffusion process using rule-quality metrics as reward signals. However, directly applying RL to the combinatorial space of graph-like rules often leads to severe sample inefficiency. To mitigate this issue,~\model~integrates a Supervised Learning (SL) pre-training strategy on subgraphs sampled from the KG meta-graph, which provides the denoising network with graph structure priors. By combining SL for structural pattern recognition with RL for quality optimization,~\model~enables the efficient generation of high-quality graph-like rules.

The contributions of this work are summarized as follows:
\begin{itemize}

\item We propose~\model, a framework that reformulates rule discovery as a conditional generative process using a diffusion model, shifting this task from discrete search to generative modeling.

\item We introduce a two-stage training strategy that employs SL pre-training to learn structure priors and RL fine-tuning to optimize the diffusion model using non-differentiable rule-quality metrics.

\item Experiments show the effectiveness of~\model~for the KGC task, while ablation studies reveal the complementary effect of graph-like rules and the correlation between their effectiveness and KG structural properties.

\end{itemize}

\section{Related Work}

\subsection{Logical Rule Learning} 

Logical rule learning is one of the main paradigms for KG reasoning, alongside embedding-based and path-based methods~\citep{ji2021survey}. Compared with embedding-based and path-based methods, rule-based reasoning provides higher interpretability by deriving conclusions through explicit logical formulas. Rule-based reasoning applies predefined logical axioms and inference rules to derive new factual knowledge from existing KGs. Owing to this interpretability, substantial research has focused on developing methods for efficiently discovering high-quality logical rules from KGs.

Logical rule learning has evolved through three methodological stages. Early approaches primarily relied on inductive logic programming and association rule mining. Foundational systems such as FOIL~\citep{Quinlan1990LearningLD} and AMIE~\citep{AMIE2013luis} generate Horn clauses by generalizing from specific instances, while other approaches are tailored to large graphs~\citep{fan2015association}. Subsequent research explored path-based methods, including AnyBURL~\cite{AnyBURL2023}, which samples and evaluates relational paths from KGs to construct logical rules. Related meta-path learning methods further address schema-complex heterogeneous information networks by learning schema-level path patterns without exhaustive path-instance enumeration~\cite{liu2023inductive}. While these methods offer interpretability, their dependence on paths or schema-level sequences limits their ability to capture non-linear graph-like rule structures.

Neural-symbolic integration paradigms were introduced to address the inefficiencies of discrete search. Early work in this direction, such as Neural Theorem Provers (NTPs)~\citep{rocktäschel2017endtoenddifferentiableproving}, introduced differentiable reasoning mechanisms. To reduce the computational demands of these approaches, subsequent research introduced optimizations including adaptive path selection~\citep{minervini2018neuraltheoremprovingscale} and dynamic rule subsetting~\citep{minervini2020learningreasoningstrategiesendtoend}. Despite these improvements, scalability challenges continued to hinder practical deployment. Consequently, recent research has shifted toward neural-logic frameworks that reformulate rule learning as a continuous optimization problem. Neural-LP~\citep{yang2017differentiable} introduced an attention-based controller for sequential rule construction, while hybrid architectures such as RNNLogic~\citep{qu2020rnnlogic} jointly learn a rule generator and a reasoning predictor for the iterative refinement of rule embeddings. NCRL~\citep{cheng2023NCRL} employs a compositional learning strategy, combining rule components through neural representation learning to improve expressiveness.

Despite these advances, existing rule learning methods still focus mainly on chain-like structures, thereby overlooking graph-like structures that can model complex relational patterns in KGs. Moreover, generating graph-like rules is computationally challenging due to the exponential growth of the rule space. To bridge this gap, we leverage diffusion models to generate graph-like rules that capture complex relational patterns for KG reasoning.

\subsection{Diffusion Models}

Diffusion models have demonstrated generative capabilities across a range of domains~\citep{yang2023diffusion}. They have also been applied to graph-structured data generation, including molecular design, where they model complex constrained output spaces~\citep{vignac2022digress,huang2024learning,huang2023conditional,watson2023novo}. This ability to model structured patterns under combinatorial constraints makes diffusion models suitable for KG reasoning tasks, which involve discovering valid relational paths within structured graphs.

Recent applications of diffusion models to KG reasoning can be grouped into two trajectories. The first strand operates at the instance level, reformulating KG reasoning as a conditional generation task in which denoising processes progressively recover target entities or facts~\citep{long2024fact, long2024}. While this paradigm learns the distribution of valid facts and achieves competitive performance, it remains confined to instance-level prediction and does not explicitly capture schema-level logical patterns in KGs. The second trajectory applies diffusion processes in continuous latent spaces to learn fact distributions and model relational uncertainty~\citep{cai2024predicting}. This line of work improves representational flexibility and can model complex relational patterns. However, because these methods focus on refining entity representations rather than discovering schema structures, they provide limited interpretability.

While both paradigms demonstrate the adaptability of diffusion models to KG reasoning, neither directly generates explicit rules beyond the instance level, and therefore neither explicitly captures schema-level logical patterns. However, directly applying diffusion models to rule generation is challenging because standard reconstruction losses are not aligned with discrete rule-quality metrics. To address this issue, we employ RL to optimize the diffusion model using rule-quality metrics as reward signals, enabling the diffusion process to be applied to graph-like rule generation.

\section{Preliminaries}

This section formalizes the key concepts used in this work. We begin by reviewing the basic structure of KGs and logical rules. We then introduce the concept of graph-like rules. Finally, we present four quantitative metrics for evaluating rule quality.

\noindent \textbf{Definition 1 (Knowledge Graph, KG).} A KG $\mathcal{G}$ is formally defined as a relation-labeled directed graph $\mathcal{G}=(\mathcal{E}, \mathcal{R}, \mathcal{F})$, where $\mathcal{E}$ denotes the entity set, $\mathcal{R}$ denotes the relation set, and $\mathcal{F}$ denotes the fact set. Specifically, $\mathcal{F} = \{(e_h,r,e_t)|e_h,e_t \in \mathcal{E}, r\in \mathcal{R}\}$ is the set of factual triples.

\noindent \textbf{Definition 2 (Logical Rule).} A logical rule $\rho$ represents a logical implication between a head predicate and a body composed of conjunctive predicates. It has the following general form:
\begin{equation}
  \rho: \rho_h \leftarrow \rho_b,  
\end{equation}
where $\rho_h$ denotes the rule head, typically formulated as a single atom such as $r_h(e_i, e_j)$, and $\rho_b$ denotes the rule body, which is formed by a conjunction of atoms. Semantically, the rule asserts that if the body $\rho_b$ is true, then the head $\rho_h$ is also true.

\begin{figure}[t]
\centering
\begin{subfigure}{0.48\columnwidth}
    \centering
    \includegraphics[width=\linewidth]{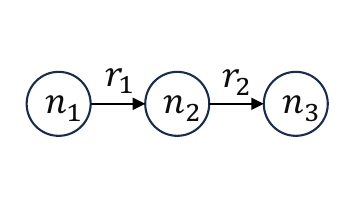}
    \caption{Linear Rule Body}
    \label{fig:pic3a}
\end{subfigure}
\hfill
\begin{subfigure}{0.48\columnwidth}
    \centering
    \includegraphics[width=\linewidth]{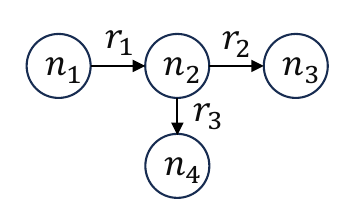}
    \caption{Branching Graph Rule Body}
    \label{fig:pic3b}
\end{subfigure}

\vspace{0.2cm} 

\begin{subfigure}{0.48\columnwidth}
    \centering
    \includegraphics[width=\linewidth]{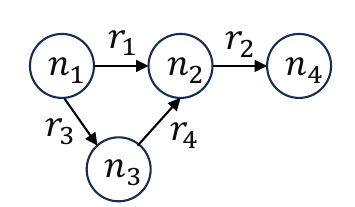}
    \caption{Edge-cycle Graph Rule Body}
    \label{fig:pic3c}
\end{subfigure}
\hfill
\begin{subfigure}{0.48\columnwidth}
    \centering
    \includegraphics[width=\linewidth]{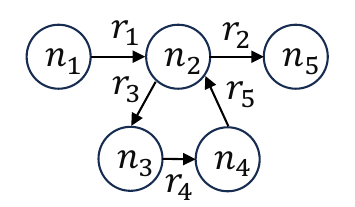}
    \caption{Node-cycle Graph Rule Body}
    \label{fig:pic3d}
\end{subfigure}

\caption{Four Basic Types of Graph-like Rule Bodies.}
\label{fig3:metagraph_type}
\end{figure}

\label{Def: graph-like rules}
\noindent \textbf{Definition 3 (Graph-like Rule).} A graph-like rule is a logical rule whose body $\rho_b$ forms a directed graph. As illustrated in Fig.~\ref{fig3:metagraph_type}, we categorize such rules into four basic types based on the connectivity pattern of the rule body: linear rule (Fig.~\ref{fig:pic3a}), which represents sequential dependencies; branching structure (Fig.~\ref{fig:pic3b}), which captures conjunctive paths; edge-cycling structure (Fig.~\ref{fig:pic3c}), which represents relation-centric cycles; and node-cycling structure (Fig.~\ref{fig:pic3d}), which models node-centric cycles. These basic structures can be combined to form more complex hybrid rule patterns.

\noindent \textbf{Definition 4 (Support).} The support of a logical rule $\rho$, denoted as $Supp_{\rho}^{\mathcal{G}}$, is the number of distinct entity pairs $(e_i, e_j)$ in $\mathcal{G}$ for which both $\rho_b$ and $\rho_h$ hold. It is defined as follows:
\begin{equation}
Supp_{\rho}^{\mathcal{G}} := \#\{(e_i, e_j) \in \mathcal{G} : \mathbb{I}_{\rho_b}(e_i, e_j) \land \rho_h(e_i, e_j)\}.
\label{equ_support}
\end{equation}

\noindent \textbf{Definition 5 (Coverage).} The coverage of a rule $\rho$ quantifies its prevalence among facts of the head relation. It is defined as the proportion of entity pairs connected by $\rho_h$ that also satisfy $\rho_b$. It is defined as follows:
\begin{equation}
Cov_{\rho}^{\mathcal{G}} := \frac{Supp_{\rho}^{\mathcal{G}}}{\#\{(e_i, e_j) \in \mathcal{G} : \rho_h(e_i, e_j)\}}.
\label{equ_cover}
\end{equation}

\noindent \textbf{Definition 6 (Confidence).} The confidence of a rule $\rho$ measures the proportion of body groundings that also support the rule head. It is defined as follows:
\begin{equation}
Conf_{\rho}^{\mathcal{G}} := \frac{Supp_{\rho}^{\mathcal{G}}}{\#\{(e_i, e_j) \in \mathcal{G} : \mathbb{I}_{\rho_b}(e_i, e_j)\}}.
\label{equ_conf}
\end{equation}

\noindent \textbf{Definition 7 (PCA Confidence).} PCA Confidence quantifies the precision of a rule $\rho$ under the Partial Completeness Assumption (PCA)~\cite{AMIE2013luis}. It restricts the denominator to body groundings whose subject entity has at least one observed fact under the head relation, and is defined as follows:
\begin{equation}
PCA\text{-}Conf_{\rho}^{\mathcal{G}} := \frac{Supp_{\rho}^{\mathcal{G}}}{\#\{{(e_i, e_j') \in \mathcal{G} : \mathbb{I}_{\rho_b}(e_i, e_j') \land \rho_h(e_i, e_j')\}}}.
\label{equ_pcaconf}
\end{equation}

\begin{figure*}[t]
    \centering
    \includegraphics[width=\linewidth]{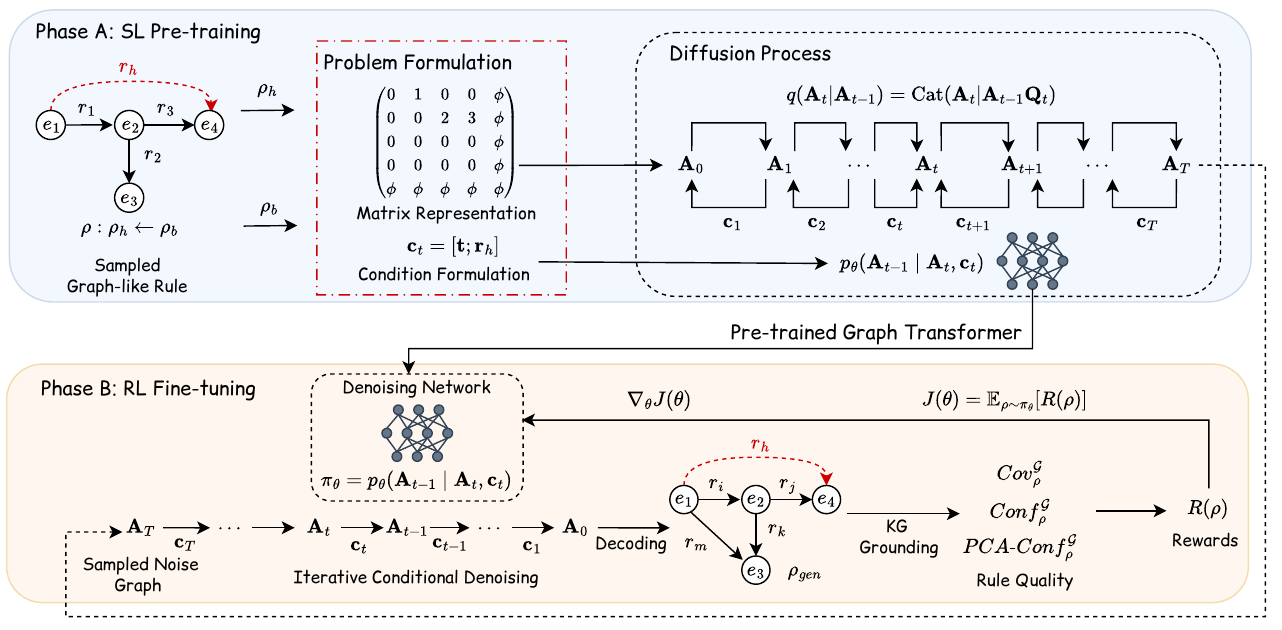}
    \caption{The~\model~framework.~\model~formulates graph-like rule discovery as a conditional discrete diffusion process over the rule-body adjacency matrices. Phase A (SL pre-training) learns graph-structure priors by corrupting subgraphs sampled from the KG meta-graph with categorical noise and training a Graph Transformer to reverse the diffusion process conditioned on the target relation. \textbf{Phase B (RL fine-tuning)} treats the denoising process as a stochastic policy, in which a rule is generated from a randomly initialized noisy graph through reverse diffusion, and the RL reward is computed from rule-quality metrics based on grounded rule instances in the KG. After RL fine-tuning,~\model~performs inference by reverse diffusion from sampled noise to generate top-ranked graph-like rules.}

    \label{fig2_framework}
\end{figure*}

\section{Methods}

This section presents~\model, a framework that formulates graph-like rule discovery as a conditional discrete diffusion process over rule-body adjacency matrices. By leveraging a diffusion model,~\model learns a distribution over KG rules conditioned on the target relation, thereby reducing dependence on explicit combinatorial search. \model operates in two stages: it is first pre-trained via SL to capture structural priors from subgraphs sampled from KG meta-graph and then fine-tuned with RL to optimize non-differentiable rule-quality metrics. An overview of the pipeline is provided in Fig.~\ref{fig2_framework}. The following sections detail the discrete diffusion framework, the two-stage training process, and the inference pipeline.

\subsection{Discrete Diffusion Framework}
\label{sec:diffusion_framework}
\model~adopts a discrete diffusion paradigm to model the symbolic structure of graph-like rules. While continuous latent representations may introduce ambiguity when reconstructing such structures, operating in the discrete space preserves their structural constraints. Accordingly,~\model~reformulates rule discovery as a conditional discrete diffusion process. The following subsections detail this framework, beginning with the discrete state formulation and followed by formal descriptions of the forward and reverse diffusion processes.

\subsubsection{Discrete State Formulation}
\label{sec:discrete_state}

The process is conditioned on the rule head $\rho_h$, which provides the semantic target, while the rule body $\rho_b$ is modeled as the adjacency matrix to be generated.

\noindent \textbf{Discrete Matrix Representation.} 
The rule body of graph-like rules is a directed graph, as defined in Def.~\ref{Def: graph-like rules}, and can therefore be represented as a matrix. Formally, the rule body adjacency matrix $\mathbf{A}$ is defined as:
\begin{equation}
    \mathbf{A} \in \{0, 1, \dots, |\mathcal{R}|, \emptyset\}^{S \times S}.
\end{equation}
Here, $S$ denotes the maximum number of allowed nodes and $\mathcal{R}$ denotes the relation set of the KG. Each entry $\mathbf{A}[i,j] \in \{0, 1, \dots, |\mathcal{R}|,\emptyset\}$ specifies the relation type from node $i$ to node $j$, with a value of $0$ indicating no relation and $\emptyset$ indicating padding. To accommodate variable-length rules, unused rows and columns are padded with a null token $\emptyset$, enabling fixed-dimensional processing for batch training.

\noindent \textbf{Conditional Formulation.}
Generation is conditioned by two factors: the target head relation $r_h$ and the diffusion timestep $t$. The relation $r_h$ is embedded into a static vector $\mathbf{r}_h \in \mathbb{R}^d$, while $t$ is encoded via sinusoidal embeddings into a time-varying vector $\mathbf{t} \in \mathbb{R}^d$. These embeddings are concatenated to form a unified conditioning vector $\mathbf{c}_t$:
\begin{equation} 
\label{eq:condition_vec} 
\mathbf{c}_t = [\mathbf{t} \mathbin{;} \mathbf{r}_h] \in \mathbb{R}^{2d}.
\end{equation} 
The learning objective is to approximate the conditional distribution $p_\theta(\mathbf{A} \mid \mathbf{c}_t)$, so that the generated rule body is conditioned on $r_h$ while respecting structural constraints at each diffusion step.

\subsubsection{Forward Diffusion Process.}
The forward diffusion process progressively corrupts a clean rule-body adjacency matrix $\mathbf{A_0}$ into noise over $T$ timesteps. Following the D3PM framework~\cite{austin2021structured}, the forward process is specified by a sequence of categorical transition matrices $\{\mathbf{Q}_t\}_{t=1}^T$. At each timestep $t$, the diffusion process factorizes across all edge positions. Each entry of $\mathbf A_t$ is represented as a $K$-dimensional one-hot categorical variable and evolves independently according to a categorical Markov transition:
\begin{equation}
q(\mathbf{A}_t \mid \mathbf{A}_{t-1}) = \textit{Cat}(\mathbf{A}_t \mid \mathbf{A}_{t-1} \mathbf{Q}_t).
\end{equation}
Here, $\textit{Cat}(\mathbf{A}_t \mid \mathbf{P})$ denotes a collection of independent categorical distributions over all edge entries, where each probability vector is given by the corresponding row of $\mathbf{P}$, and $\mathbf{Q}_t \in \mathbb{R}^{K \times K}$ is a row-stochastic transition matrix over the
$K = |\mathcal{R}| + 1$ edge types, including the no edge category. In this work, we adopt a uniform corruption scheme defined as follows:
\begin{equation}
\mathbf{Q}_t = (1 - \beta_t)\mathbf{I} + \beta_t \frac{\mathbf{1}\mathbf{1}^\top}{K},
\end{equation}
where $\beta_t \in [0,1]$ represents the noise schedule. Here, $\mathbf{1} \in \mathbb{R}^{K}$ denotes an all-ones column vector, and $\mathbf{1}\mathbf{1}^\top \in \mathbb{R}^{K \times K}$ defines a uniform categorical transition. Under this construction, each edge is left unchanged with probability $1-\beta_t$ and is resampled uniformly from all $K$ edge types with probability $\beta_t$. Using the Markov property, the marginal distribution at timestep $t$ admits a closed form as follows:
\begin{equation}
q(\mathbf{A}_t \mid \mathbf{A}_0)= \mathrm{Cat}\!\left(\mathbf{A}_t \;\mid\; \mathbf{A}_0\mathbf{Q}^{(t)}\right),
\quad
\mathbf{Q}^{(t)} = \prod_{i=1}^{t} \mathbf{Q}_i,
\end{equation}
which converges to a uniform categorical distribution as $t$ increases, thereby removing information about the original rule structure.

\subsubsection{Reverse Diffusion Process.}
This process iteratively denoises a noisy rule structure to reconstruct a clean rule-body adjacency matrix $\mathbf{A}_0$ from an initial state $\mathbf{A}_T$. The reverse transition $p_\theta(\mathbf{A}_{t-1} \mid \mathbf{A}_t, \mathbf{c}_t)$ is parameterized by a Graph Transformer $f_\theta$, which takes the noisy adjacency matrix $\mathbf{A}_t$ and the conditioning vector $\mathbf{c}_t$ (Eq.~\ref{eq:condition_vec}) as input and outputs categorical logits for all edge types.

To condition generation on the current noise level and target relation, $\mathbf{c}_t$ is injected into each network layer via Feature-wise Linear Modulation (FiLM)~\citep{perez2018film}. For the $l$-th layer, the node representation $\mathbf{H}^{(l)}$ is modulated as follows:

\begin{equation}
    \mathbf{H}^{(l)} = \boldsymbol{\gamma}(\mathbf{c}_t) \odot \text{LayerNorm}(\mathbf{H}^{(l)}) + \boldsymbol{\beta}(\mathbf{c}_t),
\end{equation}
where $\odot$ denotes element-wise multiplication, $\boldsymbol{\gamma}(\cdot)$ and $\boldsymbol{\beta}(\cdot)$ are learned affine projections of $\mathbf{c}_t$. This modulation enables the message-passing dynamics to adapt to both the timestep $t$ and the target relation $r_h$.

Finally, the modulated representations are projected to produce a probability distribution over the $K$ edge categories for each edge:
\begin{equation}
    p_\theta(\mathbf{A}_{t-1} \mid \mathbf{A}_t, \mathbf{c}_t) \propto \textit{Cat}\!\left(\mathbf{A}_{t-1} \;\mid\; f_\theta(\mathbf{A}_t, \mathbf{c}_t) \right).
\end{equation}
By iteratively sampling from this distribution,~\model~generates a rule body conditioned on the target relation while preserving structural constraints.

\subsection{Phase A: SL Pre-training}
\label{sec:supervised_pretraining}

To obtain a structurally informed initialization, we pre-train~\model~to reconstruct subgraphs sampled from the KG meta-graph. The subgraph training set is constructed through random-walk sampling and topological augmentation on the KG, illustrated in~\ref{appendix:data_construction}.

\noindent \textbf{Training Objective.}
The objective is to train the denoising network $f_\theta$ to recover the clean adjacency matrix $\mathbf{A}_0$ from a corrupted state. Specifically, given a noisy input $\mathbf{A}_t$ and the conditioning vector $\mathbf{c}_t$, we minimize a weighted cross-entropy loss between the predicted edge-type distribution and the clean adjacency matrix:
\begin{equation}\resizebox{0.9\linewidth}{!}{ $\displaystyle
    \mathcal{L}_{recon} = - \frac{\sum_{i,j} M_{ij} \sum_{k=0}^{K-1}\mathbb{I}(\mathbf{A}_0[i,j] = k) \log p_\theta\left(\mathbf{A}_0[i,j] = k \mid \mathbf{A}_t, \mathbf{c}_t\right)}{\sum_{i,j} M_{ij}}.$
    }
    \label{eq:ce_loss}
\end{equation}
Here, $\mathbf{M} \in \{0,1\}^{S \times S}$ is a binary mask, where $M_{ij}=0$ for padding positions and $M_{ij}=1$ otherwise. This ensures that the optimization focuses only on valid structural entries.

In addition to the reconstruction loss, we use the diffusion model to predict a scalar score $\hat{v}_{\rho_i}$ for each rule. The overall supervised pre-training objective is therefore defined as
\begin{equation}
    \mathcal{L}_{SL} = \mathcal{L}_{recon} + \lambda_{Q} \frac{1}{B} \sum_{i=1}^{B}\left(\hat{v}_{\rho_i} - v_{\rho_i}^*\right)^2,
\end{equation}
where $\lambda_Q=0.1$ is a fixed hyperparameter that balances structural reconstruction and semantic regularization. The target quality score $v_{\rho_i}^*$ is defined as the min-max normalized sum of coverage and confidence ($Cov_{\rho_i}^{\mathcal{G}} + Conf_{\rho_i}^{\mathcal{G}}$) over the training set, thereby accounting for both rule recall and precision. This auxiliary objective aligns the learned representations with rule-quality signals and provides a warm start for subsequent RL fine-tuning.

\noindent \textbf{Training Pipeline.} Each supervised pre-training iteration corresponds to a single diffusion step conditioned on the target relation. Concretely, we first sample a batch of rule structures $\{(\mathbf{A}^i_0, r_h)\}_{i=1}^B$ from the training set, where $B$ denotes the batch size. A diffusion timestep $t$ is then drawn uniformly from $\{1,\dots,T\}$, and the corresponding noisy structure $\mathbf{A}_t$ is obtained through the forward diffusion process. Conditioned on the vector $\mathbf{c}_t$,~\model~predicts both the edge-type distributions and an auxiliary quality score for the corrupted structure. The supervised loss $\mathcal{L}_{SL}$ is then computed and back-propagated to update the model parameters $\theta$.

\subsection{Phase B: RL Fine-tuning}
\label{sec:rl_finetuning}

While SL pre-training establishes the structural prior from KG subgraphs, it does not directly optimize rule-quality metrics such as coverage and confidence. Inspired by a recent study~\citep{kou2026pu}, the pre-trained model is fine-tuned using RL to address this mismatch, which frames the sequential denoising process as a policy optimization problem.

\noindent \textbf{Policy and Reward Formulation.}
We formulate the denoising process as a Markov Decision Process, where the denoising Graph Transformer acts as a stochastic policy $\pi_\theta(\mathbf{A}_{t-1} \mid \mathbf{A}_t, \mathbf{c}_t)$. The state corresponds to the noisy rule-body matrix $\mathbf{A}_t$, and the action is the sampled denoised matrix $\mathbf{A}_{t-1}$. Since intermediate noisy states lack direct rule-quality labels, the reward $R(\rho)$ is assigned only upon generating the complete rule $\rho$. To quantify rule quality, we define $R(\rho)$ as the geometric mean of coverage (Eq.~\ref{equ_cover}), confidence (Eq.~\ref{equ_conf}), and PCA confidence (Eq.~\ref{equ_pcaconf}):
\begin{equation}
    R(\rho) = \left( Cov_{\rho}^{\mathcal{G}} \cdot Conf_{\rho}^{\mathcal{G}} \cdot PCA\text{-}Conf_{\rho}^{\mathcal{G}} \right)^{1/3}.
    \label{equ:reward}
\end{equation}
This composite metric balances rule generality and predictive precision, providing a reward signal for optimization.

\noindent \textbf{Policy Optimization.}
Since the intermediate states do not have direct rule-quality labels, we formulate the objective as maximizing the expected episode-level reward $J(\theta) = \mathbb{E}_{\rho \sim \pi_\theta}[R(\rho)]$ using the REINFORCE algorithm. To reduce variance, we incorporate a baseline $b$, updated by an exponential moving average, to compute the advantage $\hat{A} = R(\rho) - b$. The optimization objective $\mathcal{L}_{RL}$ combines the policy gradient with entropy regularization to discourage premature convergence to deterministic transitions and encourage exploration:

\begin{equation}\resizebox{0.9\linewidth}{!}{$\displaystyle
    \mathcal{L}_{RL}(\theta) = - \frac{1}{B} \sum_{i=1}^B \sum_{t=1}^T \left( \hat{A}_i \log \pi_\theta(\mathbf{A}_{t-1}^{(i)} \mid \mathbf{A}_t^{(i)}, \mathbf{c}_t^{(i)}) + \lambda_H \mathcal{H}(\pi_\theta(\cdot \mid \mathbf{A}_t^{(i)},\mathbf{c}_t^{(i)})) \right)$.}
\end{equation}
Here, $\hat{A}_i$ acts as a trajectory-level weight for episode $i$, reinforcing steps that lead to higher-quality rules, while the entropy term $\mathcal{H}$ encourages exploration at each timestep.

\subsection{Inference Pipeline}
\label{sec: inference pipeline}

After training, the diffusion model is used as a conditional generator of graph-like rule bodies for a given target relation $r_h$. Inference proceeds via iterative denoising followed by structural parsing and filtering. Specifically, we initialize the process from a noisy adjacency matrix $\mathbf{A}_T$, where each entry is sampled from the $K$ edge categories. From $t = T$ to $t = 1$, the model iteratively samples
\begin{equation}
    \mathbf{A}_{t-1} \sim p_\theta(\mathbf{A}_{t-1} \mid \mathbf{A}_t, \mathbf{c}_t),
\end{equation}
where the conditioning vector $\mathbf{c}_t = [\mathbf{t} \mathbin{;} \mathbf{r}_h]$ encodes the timestep and target relation. This denoising process yields a discrete adjacency matrix $\mathbf{A}_0$, which is then parsed into a graph-like rule body.

To form a rule, the node with zero in-degree is identified as the head variable $e_h$. The node with zero out-degree that is topologically farthest from $e_h$ is designated as the tail variable $e_t$. The remaining edges compose the rule body that predicts $r_h(e_h, e_t)$. Finally, KG structural constraints are applied to filter invalid structures, retaining valid graph-like rules for downstream reasoning.
\begin{table*}[t]
    \centering
    \renewcommand\arraystretch{0.95}
\caption{KGC results of~\model~on six datasets. The best/second-best metrics are bolded/underlined. All reported results are averaged over five independent runs with different random seeds, detailed in~\ref{sec:seed_analysis}.}
    \resizebox{\linewidth}{!}{%
    \begin{tabular}{ccccccccccc}
     \toprule[1.5pt]
          & \multirow{2}{*}{Models} & \multicolumn{3}{c}{Family} & \multicolumn{3}{c}{Kinship}       & \multicolumn{3}{c}{UMLS}    \\
    &    & MRR (\%)   & HIT@1 (\%) & HIT@10 (\%) & MRR (\%)   & HIT@1 (\%) & HIT@10 (\%)& MRR (\%)   & HIT@1 (\%) & HIT@10 (\%) \\
    \hline
    \multirow{5}{*}{Embedding} 
    & TransE    & 45.21 & 22.34 & 87.45   & 31.23 & 0.93 & 83.22 & 69.86 & 52.14  & 89.19    \\
    & DistMult  & 54.17 & 36.00  & 88.45   & 35.23 & 19.22  & 74.23  & 39.04 & 25.61  & 66.15    \\
    & ComplEx   & 81.11 & 72.12  & 93.63  & 42.12 & 23.56 & 82.11 & 41.29 & 27.93  & 70.92      \\
    & RotatE    & 86.70 & 77.43  & 93.21   & \underline{66.21} & 50.24  &  \textbf{94.13} & 74.21 & 63.92  & 94.04   \\
   
    & SpeedE & 90.28 & 82.72 & 98.47 & 65.32 & \underline{53.35} & 89.55  & 70.35 & 58.62  & 89.65   \\ 
    
    \hline
    \multirow{3}{*}{GNN-based} & CompGCN    & 75.35 &58.83 &88.83 &40.30 &26.94 &75.83   &62.33 &50.85&72.76  \\
& NBFNet   & 84.16 & 76.03 & 97.71 & 49.54 & 34.43  & 81.07   & 70.74 & 64.19  & 82.48     \\
& A*Net   & 82.35 & 72.37 & 96.67 & 44.36 & 30.42 & 77.36 & 66.37 &53.63 &  78.37 \\

    \hline
    \multirow{9}{*}{Rule-based}
    & AnyBURL & \underline{93.99} & \underline{88.46}  & 94.90   & 65.93 & 52.38  & 91.45    & 68.83 & 67.40  & 72.54  \\
     & AMIE & 87.74 & 80.47 & 98.83   & 62.99 & 48.83  & 90.06    & 64.42 & 60.67  & 74.21  \\
    
    & Neural-LP & 88.12 & 79.28  & 98.53   & 30.17 & 16.60  & 59.34     & 48.38 & 33.28  & 77.47  \\
    & DRUM      & 89.03 & 82.76  & \underline{99.06}   & 33.02 & 18.11  & 67.76   & 55.18 & 35.37  & 85.15  \\
    & RNNLogic  & 86.74 & 79.40  & 95.18   & 64.78 & 49.70  & 92.26   & 75.20 & 62.40  & 92.04\\		
    & RLogic    & 88.08 & 80.81  & 97.27   & 57.70 & 43.60  & 87.08     & 71.74 & 55.32  & 93.55   \\
    & NCRL      &91.33 & 85.24 &\textbf{{99.31}}   & 64.32 & 49.39  & \underline{92.56} & 75.53 & 64.44 & 93.15     \\
    & ChatRule   & 90.62 & 85.45 &  96.84   & 32.52 & 17.51  & 68.74 & 77.51 & \underline{67.45} &  \underline{94.84}   \\

    & \model       & \textbf{95.25} & \textbf{91.54} & 98.32   & \textbf{67.18} & \textbf{54.36}  & {92.46}   &  \textbf{83.48} & \textbf{74.89}  & \textbf{96.48}  \\
    \midrule[1pt]
   
    \midrule[1pt]
    & \multirow{2}{*}{Model} & \multicolumn{3}{c}{WN-18RR} & \multicolumn{3}{c}{FB15K-237}             & \multicolumn{3}{c}{YAGO3-10}                \\
    &     & MRR (\%)   & HIT@1 (\%) & HIT@10 (\%) & MRR (\%)   & HIT@1 (\%) & HIT@10 (\%) & MRR (\%)   & HIT@1 (\%) & HIT@10 (\%)\\
    \hline
    \multirow{5}{*}{Embedding} & 
        TransE   & 23.12 & 2.23   & 52.54 & 29.40 & 18.65  & 46.84   & 36.42 & 25.78  & 58.01   \\
         & DistMult & 42.67 & 38.06  & 50.38 & 22.53 & 13.93  & 38.46   & 34.56 & 24.54  & 52.93   \\
         & ComplEx & 44.11 & 41.35  & 51.15   & 24.49 & 15.26  & 41.67   & 33.34 & 24.23  & 53.86   \\
         & RotatE   & 47.17 & 42.92  & {55.74}   & 32.40 & 22.24  & 52.48  & 49.41 & 40.76  & \textbf{67.70}  \\
         
        & SpeedE &49.31 &44.52& {57.63}& 32.71 & 22.32 & 49.57 & \underline{41.35} & \underline{33.25} & {51.35}\\
   
    \hline
    \multirow{3}{*}{GNN-based}
    & CompGCN  & 45.32 & 39.25 & 54.63 &  33.74 &22.98 &52.55 &28.43 &21.36 &47.52  \\
    & NBFNet   & \textbf{{53.87}} & \textbf{{49.50}}  & \underline{62.93} & 39.22 & 28.79  & 57.39   & 36.83 & 27.91 & 55.42     \\
    & A*Net   & \underline{51.18} & 42.82 & \textbf{64.37} & 39.88 & 30.89 & 58.13  & 36.16 & 27.36 & 55.93 \\
    \hline
    
    \multirow{9}{*}{Rule-based}
    & AnyBURL & 41.20 & 37.91  & 47.90   & \underline{41.32} & \underline{30.97}  & \underline{61.40}    & 46.14 & 39.60  & 58.52  \\
     & AMIE & 36.02 & 33.42  & 41.00   & - & -  & -    & - & -  & -  \\ 
    &Neural-LP  & 37.67 & 36.46  & 40.15  & 24.15 & 17.71  & 36.56   & -          & -           & -            \\
    & DRUM  & 36.68 & 38.41  & 41.30  & 22.89 & 17.23  & 35.75   & -          & -           & -            \\
    & RNNLogic  &46.28 & 41.40  & 53.32  & 28.59 & 20.20  & 44.62   & -          & -                     \\
    & RLogic   & 44.23 & 40.14  & 50.01  & 31.10 & 19.78  & 50.02   & 36.23 & 25.23  & 50.25   \\
    & NCRL     &  46.68 & \underline{44.64}  & 53.22& 29.76 & 20.24  & 47.46   & 38.33 & 27.87  & 53.14   \\
    & ChatRule    & 33.51 & 30.12  & 40.43& 30.91 & 22.32 &  49.88   & 44.92 & 35.41 & 62.70 \\ 

    & \model   & {45.10} & {40.22}  & 54.39 & \textbf{42.29} & \textbf{32.05}  & \textbf{62.13}  & \textbf{54.91}    & \textbf{49.78}    &  \underline{62.97}  \\         
         \bottomrule[1.5pt]
    \end{tabular}
    }
    \label{table:KGC}
    \end{table*}

\section{Experiment}
\subsection{Experimental settings}

\noindent\textbf{Tasks.} We evaluate the performance and efficiency of~\model~on KGC tasks. The objective is to predict a missing target entity $e_t$ for a given query $(e_h, r_q, ?)$. KGC tasks assess the capability of~\model to extract high-quality logical rules for KG reasoning. The full KGC pipeline is detailed in~\ref{sec:eval_process}.

\noindent\textbf{Datasets.} Our evaluation spans six benchmark datasets representing diverse scales and domains: two large-scale KGs, YAGO3-10~\citep{suchanek2007yago} and FB15K-237~\citep{toutanova2015observed}, alongside four domain-specific datasets: Family~\citep{hinton1986learning}, Kinship~\citep{kok2007statistical}, UMLS~\citep{kok2007statistical}, and WN-18RR~\citep{dettmers2018convolutional}. Detailed statistics are provided in~\ref{sec:dataset}.

\noindent\textbf{Dataset Preparation.} Each dataset is randomly split into training and test sets with a ratio of 9:1. To avoid information leakage, all test triples are strictly excluded from the training data.

\noindent\textbf{Baselines.} 
We compare~\model~against 16 representative baselines spanning three paradigms under identical experimental settings: five embedding approaches, including TransE~\citep{bordes2013transe}, DistMult~\citep{yang2014embedding}, ComplEx~\citep{trouillon2016complex}, RotatE~\citep{sun2019rotate}, and SpeedE~\citep{pavlovic2024speede}; three GNN-based methods, including CompGCN~\citep{vashishth2019composition}, NBFNet~\citep{zhu2021neural}, and A$^{*}$Net~\citep{zhu2023net}; and eight rule-based systems, including AnyBURL~\citep{AnyBURL2023}, AMIE~\citep{AMIE2013luis}, Neural-LP~\citep{yang2017differentiable}, DRUM~\citep{sadeghian2019drum}, RNNLogic~\citep{qu2020rnnlogic}, RLogic~\citep{Rlogic2022}, NCRL~\citep{cheng2023NCRL}, and ChatRule~\citep{luo2023chatrule}. Detailed descriptions of these baselines are provided in~\ref{sec:baseline_appendix}.

\noindent \textbf{Metrics.} Following standard KGC evaluation protocols, we adopt filtered ranking, where triples $(e_h,r_q,e^\star)$ with $e^\star \neq e_t$ are removed from the candidate set if they exist in the KG. We report Hit@1, Hit@10, and Mean Reciprocal Rank (MRR). 

\noindent \textbf{Rule Application.} For each relation $r$, both chain-like and graph-like rules are used to compute the final scoring matrix, $\mathbf{S}_r = (1-\alpha)\mathbf{S}^{\mathrm{chain}}_r + \alpha\mathbf{S}^{\mathrm{graph}}_r$, where the fusion weight is $\alpha = 0.1$. The detailed scoring and aggregation procedure is described in~\ref{sec:eval_process}.

\subsection{Knowledge Graph Completion Results}

\noindent \textbf{Overall Results.} As summarized in Table~\ref{table:KGC},~\model~achieves consistently strong KGC performance across six benchmarks. It outperforms most rule-based baselines in the majority of settings and remains competitive with representative embedding and GNN-based baselines, while producing explicit logical rules. On large-scale KGs such as FB15K-237 and YAGO3-10, several rule-based approaches encounter scalability limitations, whereas~\model~still achieves the best or second-best results.

\noindent \textbf{Impact of KG Structure.} We observe that graph-like rules are particularly beneficial on KGs where rich relational structure leads to ambiguous predictions under simple chain-based reasoning, such as FB15K-237 and YAGO3-10. In such settings, chain-like rules often yield ambiguous candidate rankings, while graph-like rules introduce additional conjunctive constraints that help disambiguate candidate entities. Conversely, on KGs with tightly constrained relational semantics (e.g., kinship or lexical hierarchies), chain‑like rules already capture most reliable patterns, leaving less room for improvement from graph‑like rules. A detailed quantitative analysis of KG structural properties and their relationship to performance is provided in~\ref{sec:KG_structure_analysis}.

\noindent \textbf{Case Study.} To further illustrate the differences between the learned logical rules, representative examples of chain-like and graph-like rules are presented in~\ref{appendix:rule comparision}. Graph-like rules impose stricter conjunctive constraints by requiring the joint satisfaction of multiple relational conditions. For example, the rule $\textit{isCitizenOf}(x,y) \leftarrow \textit{LivesIn}(x,y) \wedge \textit{WorksAt}(x,y)$ integrates two complementary relational signals through the shared variables $(x, y)$. In contrast, chain-like rules are characterized by a single linear relational dependency.

\subsection{Ablation Studies}
We conduct ablation studies to assess several design choices of~\model, focusing on four key questions: \textbf{(Q1) Effectiveness of Graph-like Rules:} Do graph-like rules provide complementary inferential evidence beyond chain-like rules? \textbf{(Q2) Sensitivity to Fusion Weight:} Under which fusion weights do graph-like rules provide the largest performance gains? \textbf{(Q3) Necessity of RL:} Is RL fine-tuning essential for rule discovery? \textbf{(Q4) Size of Graph-like Rules:} Is a small graph size sufficient for complex KGs?

\noindent \textbf{(Q1) Effectiveness of Graph-like Rules.} To address Q1, we investigate whether graph-like rules provide complementary inferential evidence beyond chain-like rules. We compare three configurations: \emph{Chain-only}, \emph{Graph-only}, and their \emph{Combination}, while maintaining a constant total number of rules for a controlled comparison. As shown in Table~\ref{tab:rule_type_kgc_results}, the combined configuration consistently outperforms either individual type, indicating that graph‑like rules capture relational patterns distinct from those encoded by chain‑like rules. Graph-like rules in isolation achieve lower KGC performance because their stricter constraints reduce coverage in sparse KGs, which is consistent with the sparsity of KG observations.

\begin{table}[t]
\renewcommand\arraystretch{0.95}
\centering
\caption{KGC Results of Different Rule Types}
\label{tab:rule_type_kgc_results}
\resizebox{0.9\linewidth}{!}{
\begin{tabular}{lcccccc}
\toprule[1pt]
\multirow{2}{*}{Rule-Type} & \multicolumn{3}{c}{Family} & \multicolumn{3}{c}{WN-18RR} \\
\cmidrule(lr){2-4} \cmidrule(lr){5-7}
 & MRR & Hit@1 & Hit@10 & MRR & Hit@1 & Hit@10 \\
\midrule
Chain-only & 94.79 & 91.21 & 99.01 & 38.73 & 35.12 & 45.56\\
Graph-only & 90.59 & 84.72 & 97.60 & 31.57 & 27.18 & 40.26 \\
Combined & \textbf{95.32} & \textbf{91.66} & \textbf{99.65} & \textbf{45.10} & \textbf{40.22} & \textbf{54.39} \\
\bottomrule[1pt]
\toprule[1pt]
\multirow{2}{*}{Rule-Type} & \multicolumn{3}{c}{FB15K-237} & \multicolumn{3}{c}{YAGO3-10} \\
\cmidrule(lr){2-4} \cmidrule(lr){5-7}
 & MRR & Hit@1 & Hit@10 & MRR & Hit@1 & Hit@10 \\
\midrule
Chain-only  & 41.84 & 31.82 & 61.08 & 49.56 & 44.96 & 56.78 \\
Graph-only  & 37.99 & 28.64 & 55.60 &  48.39 & 43.84 & 55.44 \\
Combined & \textbf{42.29} & \textbf{32.05} & \textbf{62.13} & \textbf{55.77} & \textbf{49.78} & \textbf{62.97} \\
\bottomrule[1pt]
\end{tabular}%
}
\end{table}

\begin{figure}[t]
    \centering
    \begin{subfigure}{0.48\linewidth}
        \centering
        \includegraphics[width=\linewidth]{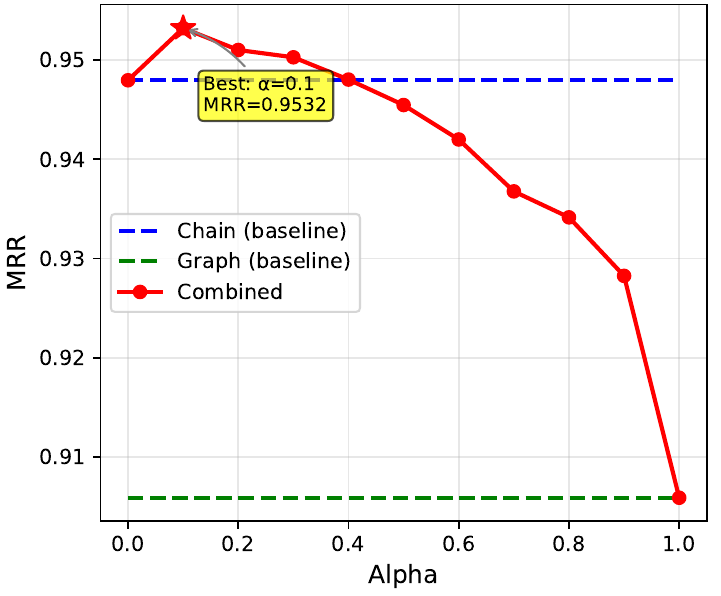}
        \caption{Family MRR}
        \label{fig:pic4_1}
    \end{subfigure}
    \hfill 
    \begin{subfigure}{0.48\linewidth}
        \centering
        \includegraphics[width=\linewidth]{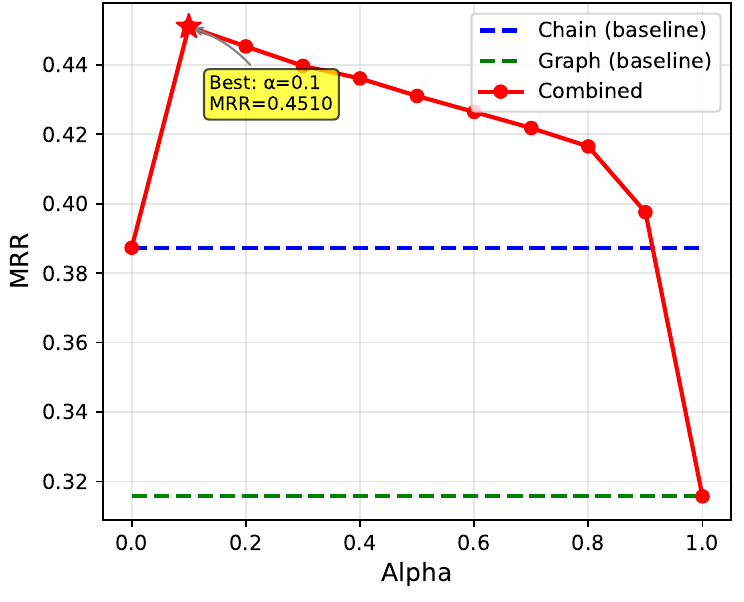}
        \caption{WN-18RR MRR}
        \label{fig:pic4_2}
    \end{subfigure}

    \vspace{10pt} 

    \begin{subfigure}{0.48\linewidth}
        \centering
        \includegraphics[width=\linewidth]{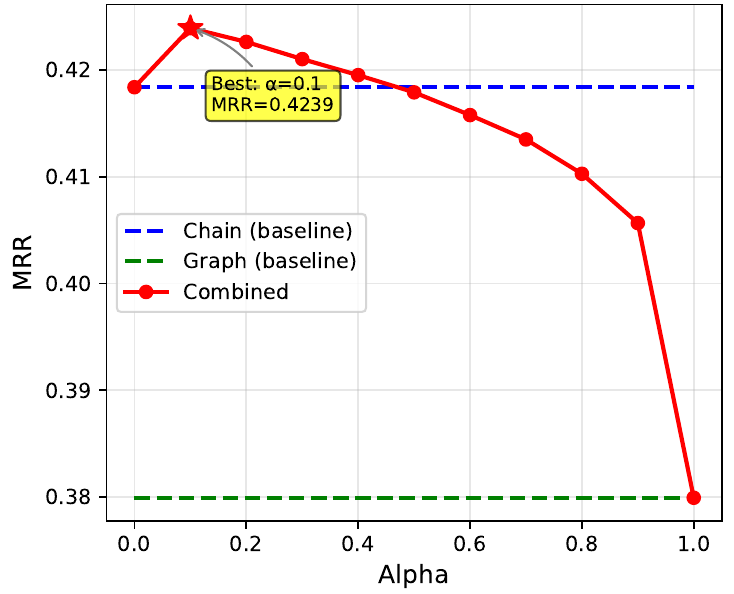}
        \caption{FB15K-237 MRR}
        \label{fig:pic4_3}
    \end{subfigure}
    \hfill
    \begin{subfigure}{0.48\linewidth}
        \centering
        \includegraphics[width=\linewidth]{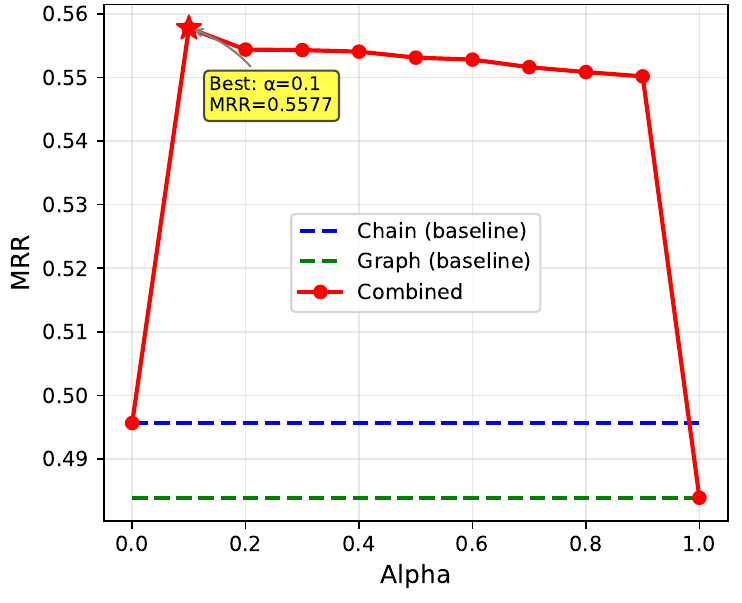}
        \caption{YAGO3-10 MRR}
        \label{fig:pic4_4}
    \end{subfigure}

    \caption{Ablation study on the fusion weight $\alpha$.}
    \label{fig:alpha_sweep}
\end{figure}

\noindent \textbf{(Q2) Sensitivity to Fusion Weight.}
To address Q2, we vary the fusion weight $\alpha$ at the scoring stage while keeping the generated chain-like and graph-like rule sets fixed. Here, $\alpha=0$ corresponds to chain-only reasoning, and $\alpha=1$ corresponds to graph-only reasoning. As shown in Fig.~\ref{fig:alpha_sweep}, adding graph-like rules improves the chain-only baseline, with the best performance typically obtained when $\alpha$ is around $0.1$--$0.2$. Larger $\alpha$ values degrade performance because graph-like rules enforce additional joins, which improve precision but reduce the number of supported groundings. These results show that the fusion score balances the coverage of chain-like rules with the stricter constraints of graph-like rules.

\begin{figure}[t]
  \centering
\includegraphics[width=\linewidth]{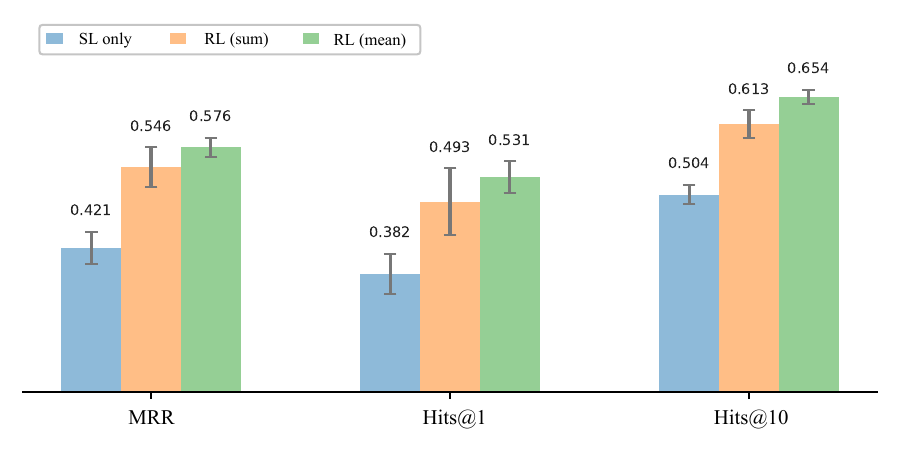}

\caption{Effect of RL fine-tuning on YAGO3-10.}
  \label{fig:RL_ablation}
\end{figure}

\noindent \textbf{(Q3) Necessity of RL.} To address Q3, we compare~\model~trained with only supervised pre-training against two RL-fine-tuned variants on YAGO3-10: one using a geometric-mean reward and the other using the sum of rule-quality metrics. As shown in Fig.~\ref{fig:RL_ablation}, RL fine-tuning yields consistent improvements of approximately 10\% across all metrics, indicating that direct optimization toward rule quality helps align generation with reasoning objectives. Moreover, the geometric-mean reward outperforms metric summation, suggesting that balanced multi-criteria optimization provides a more effective training signal in this setting.

\noindent \textbf{(Q4) Size of Graph-like Rules.} To address Q4, we conduct KGC experiments with different graph-size limits $S$. As shown in Table~\ref{tab:graph_size}, $S=6$ consistently achieves the best MRR across all datasets. This result is consistent with the compositional nature of logical rules: longer rules often provide limited additional utility because they can be expressed by composing shorter rules. Moreover, rules operate at the schema level and are therefore not related to instance-level neighbor density. Consequently, a small $S$ can improve efficiency while preserving sufficient expressiveness for the rule patterns considered in this work.

\begin{table}[t]
\centering
\caption{KGC MRR under different graph-size limits $S$.}
\label{tab:graph_size}
\begin{tabular}{cccc}
\toprule[1.5pt]
$S$ & WN-18RR & FB15K-237 & YAGO3-10 \\
\midrule
5 & 0.420          & 0.381          & 0.518 \\
6 & \textbf{0.462} & \textbf{0.411} & \textbf{0.548} \\
7 & 0.452          & 0.397          & 0.519 \\
\bottomrule[1.5pt]
\end{tabular}
\end{table}

\subsection{Efficiency Analysis}
\label{sec:efficiency_analysis}

We assess the efficiency of~\model~from both theoretical and empirical perspectives. In SL pre-training, the cost is dominated by the Graph Transformer, with per-epoch complexity $O(|\mathcal{D}_{pre}|S^2)$. In RL fine-tuning, complete denoising trajectories are sampled for reward computation, giving complexity $O(N_{RL}TS^2)$ for $N_{RL}$ sampled rules and $T$ diffusion steps. Since graph-like rule bodies are compact ($S\le6$), the quadratic dependence on $S$ introduces limited overhead. Rule generation is performed offline for each relation; therefore, the online cost for each query is independent of the denoising trajectory length $T$ and mainly comes from sparse-matrix lookup and aggregation over retained rule groundings. Thus, the denoising cost affects offline rule construction rather than KGC inference.

Empirically, as reported in Table~\ref{tab:cost_efficiency},~\model~has moderate training cost and memory usage. On YAGO3-10, SL pre-training and RL fine-tuning require 1.34 s and 18.76 s per epoch, respectively. Peak GPU memory is 11.41 GB on FB15K-237 and 5.52 GB on YAGO3-10. Although sampling on large KGs is more time-consuming, it is an offline cost incurred once per relation. After rule generation, inference is performed by sparse-matrix grounding and aggregation, supporting rule application on large-scale KGs. 


    

\begin{table}[t]
    \centering
    \caption{Mean computational cost over 5 independent runs. Time is reported in seconds (s) and peak memory in GB.}
    \label{tab:cost_efficiency}
    \renewcommand\arraystretch{1.15}
\resizebox{\linewidth}{!}{
    \begin{tabular}{lcccccc}
        \toprule[1.5pt]
        Dataset & Sampling (s) & SL/epoch (s) & RL/epoch (s) & Gen/100 (s) & Total Time (s) & Peak Mem (GB) \\
        \midrule
        Family    & 4.23    & 0.74 & 1.35  & 1.10 & 58.91   & 1.81  \\
        Kinship   & 4.87    & 0.14 & 0.46  & 0.42 & 223.67  & 3.57  \\
        UMLS      & 10.11   & 1.67 & 2.89  & 3.15 & 125.59  & 6.06  \\
        WN-18RR    & 5.84    & 0.21 & 2.57  & 4.44 & 93.99   & 1.54  \\
        FB15K-237 & 3079.35 & 8.61 & 4.96  & 0.99 & 3699.88 & 11.41 \\
        YAGO3-10  & 4479.54 & 1.34 & 18.76 & 1.03 & 7279.68 & 5.52  \\
        \bottomrule[1.5pt]
    \end{tabular}
    }
\end{table}

\section{Conclusion}

This paper introduces~\model, a framework that reformulates rule discovery as a discrete diffusion process conditioned on the target relation. By shifting from iterative search to generative modeling,~\model~reduces the dependence on exhaustive search in the combinatorial rule space. By combining SL pre-training for structural priors with RL fine-tuning that directly optimizes rule quality metrics,~\model~generates graph-like rules for KG reasoning. Experiments show that the rules generated by~\model~support KGC across six benchmark datasets. Furthermore, ablation studies show the complementary effects of graph-like rules and chain-like rules. Future research will explore generating rules directly from underlying KG distributions to mitigate sampling bias and developing structural evaluation metrics that provide more precise guidance for the diffusion process.

\section{Acknowledgements}
This study was supported by the National Natural Science Foundation of China (Nos. 72571284, 72421002, 62273352), the Hunan Provincial Fund for Distinguished Young Scholars (No. 2025JJ20073), the Science and Technology Innovation Program of Hunan Province (No. 2023RC3009), the Innovation Research Foundation of National University of Defense Technology (No. JS24-05), the Major Science and Technology Projects in Changsha, China (No. kq2301008), the Beijing Natural Science Foundation (No. L257007), the Beijing Major Science and the Technology Project (No. Z251100008425006).


\bibliographystyle{ACM-Reference-Format}
\bibliography{references_new}


\begin{thebibliography}{50}


\ifx \showCODEN    \undefined \def \showCODEN     #1{\unskip}     \fi
\ifx \showISBNx    \undefined \def \showISBNx     #1{\unskip}     \fi
\ifx \showISBNxiii \undefined \def \showISBNxiii  #1{\unskip}     \fi
\ifx \showISSN     \undefined \def \showISSN      #1{\unskip}     \fi
\ifx \showLCCN     \undefined \def \showLCCN      #1{\unskip}     \fi
\ifx \shownote     \undefined \def \shownote      #1{#1}          \fi
\ifx \showarticletitle \undefined \def \showarticletitle #1{#1}   \fi
\ifx \showURL      \undefined \def \showURL       {\relax}        \fi
\providecommand\bibfield[2]{#2}
\providecommand\bibinfo[2]{#2}
\providecommand\natexlab[1]{#1}
\providecommand\showeprint[2][]{arXiv:#2}

\bibitem[Austin et~al\mbox{.}(2021)]%
        {austin2021structured}
\bibfield{author}{\bibinfo{person}{Jacob Austin}, \bibinfo{person}{Daniel~D.
  Johnson}, \bibinfo{person}{Jonathan Ho}, \bibinfo{person}{Daniel Tarlow},
  {and} \bibinfo{person}{Rianne van~den Berg}.}
  \bibinfo{year}{2021}\natexlab{}.
\newblock \showarticletitle{Structured Denoising Diffusion Models in Discrete
  State-Spaces}. In \bibinfo{booktitle}{\emph{Advances in Neural Information
  Processing Systems ({NeurIPS})}}, Vol.~\bibinfo{volume}{34}.
  \bibinfo{pages}{17981--17993}.
\newblock


\bibitem[Barannikov et~al\mbox{.}(2021)]%
        {barannikov2021manifold}
\bibfield{author}{\bibinfo{person}{Serguei Barannikov}, \bibinfo{person}{Ilya
  Trofimov}, \bibinfo{person}{Grigorii Sotnikov}, \bibinfo{person}{Ekaterina
  Trimbach}, \bibinfo{person}{Alexander Korotin}, \bibinfo{person}{Alexander
  Filippov}, {and} \bibinfo{person}{Evgeny Burnaev}.}
  \bibinfo{year}{2021}\natexlab{}.
\newblock \showarticletitle{Manifold Topology Divergence: A Framework for
  Comparing Data Manifolds}. In \bibinfo{booktitle}{\emph{Advances in Neural
  Information Processing Systems ({NeurIPS})}}, Vol.~\bibinfo{volume}{34}.
  \bibinfo{pages}{7294--7305}.
\newblock


\bibitem[Bordes et~al\mbox{.}(2013)]%
        {bordes2013transe}
\bibfield{author}{\bibinfo{person}{Antoine Bordes}, \bibinfo{person}{Nicolas
  Usunier}, \bibinfo{person}{Alberto Garcia-Duran}, \bibinfo{person}{Jason
  Weston}, {and} \bibinfo{person}{Oksana Yakhnenko}.}
  \bibinfo{year}{2013}\natexlab{}.
\newblock \showarticletitle{Translating Embeddings for Modeling
  Multi-Relational Data}. In \bibinfo{booktitle}{\emph{Advances in Neural
  Information Processing Systems ({NeurIPS})}}, Vol.~\bibinfo{volume}{26}.
  \bibinfo{pages}{2787--2795}.
\newblock


\bibitem[Cai et~al\mbox{.}(2024)]%
        {cai2024predicting}
\bibfield{author}{\bibinfo{person}{Yuxiang Cai}, \bibinfo{person}{Qiao Liu},
  \bibinfo{person}{Yanglei Gan}, \bibinfo{person}{Changlin Li},
  \bibinfo{person}{Xueyi Liu}, \bibinfo{person}{Run Lin}, \bibinfo{person}{Da
  Luo}, {and} \bibinfo{person}{JiayeYang JiayeYang}.}
  \bibinfo{year}{2024}\natexlab{}.
\newblock \showarticletitle{Predicting the Unpredictable: Uncertainty-Aware
  Reasoning over Temporal Knowledge Graphs via Diffusion Process}. In
  \bibinfo{booktitle}{\emph{Findings of the Association for Computational
  Linguistics: {ACL} 2024 ({ACL} Findings)}}. \bibinfo{publisher}{Association
  for Computational Linguistics}, \bibinfo{pages}{5766--5778}.
\newblock


\bibitem[Chen et~al\mbox{.}(2026)]%
        {chen2026logiconbench}
\bibfield{author}{\bibinfo{person}{Zheng Chen}, \bibinfo{person}{Chuan Zhou},
  \bibinfo{person}{Fengxiang Cheng}, \bibinfo{person}{Yip~Tin Po},
  \bibinfo{person}{Fenrong Liu}, \bibinfo{person}{Yisen Wang},
  \bibinfo{person}{Jiajun Chai}, \bibinfo{person}{Xiaohan Wang},
  \bibinfo{person}{Guojun Yin}, \bibinfo{person}{Wei Lin}, \bibinfo{person}{Bo
  Li}, \bibinfo{person}{Haoxuan Li}, {and} \bibinfo{person}{Zhouchen Lin}.}
  \bibinfo{year}{2026}\natexlab{}.
\newblock \showarticletitle{Logi{C}on{B}ench: Benchmarking Logical
  Consistencies of {LLM}s}. In \bibinfo{booktitle}{\emph{The Fourteenth
  International Conference on Learning Representations}}.
\newblock


\bibitem[Cheng et~al\mbox{.}(2025)]%
        {cheng2025empowering}
\bibfield{author}{\bibinfo{person}{Fengxiang Cheng}, \bibinfo{person}{Haoxuan
  Li}, \bibinfo{person}{Fenrong Liu}, \bibinfo{person}{Robert Van~Rooij},
  \bibinfo{person}{Kun Zhang}, {and} \bibinfo{person}{Zhouchen Lin}.}
  \bibinfo{year}{2025}\natexlab{}.
\newblock \showarticletitle{Empowering LLMs with logical reasoning: a
  comprehensive survey}. In \bibinfo{booktitle}{\emph{Proceedings of the
  Thirty-Fourth International Joint Conference on Artificial Intelligence}}.
  \bibinfo{pages}{10400--10408}.
\newblock


\bibitem[Cheng et~al\mbox{.}(2023)]%
        {cheng2023NCRL}
\bibfield{author}{\bibinfo{person}{Kewei Cheng}, \bibinfo{person}{Nesreen~K.
  Ahmed}, {and} \bibinfo{person}{Yizhou Sun}.} \bibinfo{year}{2023}\natexlab{}.
\newblock \showarticletitle{Neural Compositional Rule Learning for Knowledge
  Graph Reasoning}. In \bibinfo{booktitle}{\emph{Proceedings of the
  International Conference on Learning Representations ({ICLR})}}.
\newblock


\bibitem[Cheng et~al\mbox{.}(2022)]%
        {Rlogic2022}
\bibfield{author}{\bibinfo{person}{Kewei Cheng}, \bibinfo{person}{Jiahao Liu},
  \bibinfo{person}{Wei Wang}, {and} \bibinfo{person}{Yizhou Sun}.}
  \bibinfo{year}{2022}\natexlab{}.
\newblock \showarticletitle{{RLogic}: Recursive Logical Rule Learning from
  Knowledge Graphs}. In \bibinfo{booktitle}{\emph{Proceedings of the 28th ACM
  SIGKDD Conference on Knowledge Discovery and Data Mining ({KDD})}}.
  \bibinfo{publisher}{ACM}, \bibinfo{pages}{179--189}.
\newblock


\bibitem[Dettmers et~al\mbox{.}(2018)]%
        {dettmers2018convolutional}
\bibfield{author}{\bibinfo{person}{Tim Dettmers}, \bibinfo{person}{Pasquale
  Minervini}, \bibinfo{person}{Pontus Stenetorp}, {and}
  \bibinfo{person}{Sebastian Riedel}.} \bibinfo{year}{2018}\natexlab{}.
\newblock \showarticletitle{Convolutional 2D Knowledge Graph Embeddings}. In
  \bibinfo{booktitle}{\emph{Proceedings of the AAAI Conference on Artificial
  Intelligence ({AAAI})}}, Vol.~\bibinfo{volume}{32}.
\newblock


\bibitem[Fan et~al\mbox{.}(2015)]%
        {fan2015association}
\bibfield{author}{\bibinfo{person}{Wenfei Fan}, \bibinfo{person}{Xin Wang},
  \bibinfo{person}{Yinghui Wu}, {and} \bibinfo{person}{Jingbo Xu}.}
  \bibinfo{year}{2015}\natexlab{}.
\newblock \showarticletitle{Association Rules with Graph Patterns}.
\newblock \bibinfo{journal}{\emph{Proceedings of the VLDB Endowment ({PVLDB})}}
  \bibinfo{volume}{8}, \bibinfo{number}{12} (\bibinfo{year}{2015}),
  \bibinfo{pages}{1502--1513}.
\newblock


\bibitem[Gal{\'a}rraga et~al\mbox{.}(2013)]%
        {AMIE2013luis}
\bibfield{author}{\bibinfo{person}{Luis~Antonio Gal{\'a}rraga},
  \bibinfo{person}{Christina Teflioudi}, \bibinfo{person}{Katja Hose}, {and}
  \bibinfo{person}{Fabian Suchanek}.} \bibinfo{year}{2013}\natexlab{}.
\newblock \showarticletitle{{AMIE}: Association Rule Mining under Incomplete
  Evidence in Ontological Knowledge Bases}. In
  \bibinfo{booktitle}{\emph{Proceedings of the 22nd International Conference on
  World Wide Web ({WWW})}}. \bibinfo{publisher}{ACM},
  \bibinfo{pages}{413--422}.
\newblock


\bibitem[Hinton(1986)]%
        {hinton1986learning}
\bibfield{author}{\bibinfo{person}{Geoffrey~E. Hinton}.}
  \bibinfo{year}{1986}\natexlab{}.
\newblock \showarticletitle{Learning Distributed Representations of Concepts}.
  In \bibinfo{booktitle}{\emph{Proceedings of the Eighth Annual Conference of
  the Cognitive Science Society}}. \bibinfo{publisher}{Lawrence Erlbaum
  Associates}, \bibinfo{pages}{1--12}.
\newblock


\bibitem[Huang et~al\mbox{.}(2023)]%
        {huang2023conditional}
\bibfield{author}{\bibinfo{person}{Han Huang}, \bibinfo{person}{Leilei Sun},
  \bibinfo{person}{Bowen Du}, {and} \bibinfo{person}{Weifeng Lv}.}
  \bibinfo{year}{2023}\natexlab{}.
\newblock \showarticletitle{Conditional diffusion based on discrete graph
  structures for molecular graph generation}. In
  \bibinfo{booktitle}{\emph{Proceedings of the AAAI Conference on Artificial
  Intelligence}}, Vol.~\bibinfo{volume}{37}. \bibinfo{pages}{4302--4311}.
\newblock


\bibitem[Huang et~al\mbox{.}(2024)]%
        {huang2024learning}
\bibfield{author}{\bibinfo{person}{Han Huang}, \bibinfo{person}{Leilei Sun},
  \bibinfo{person}{Bowen Du}, {and} \bibinfo{person}{Weifeng Lv}.}
  \bibinfo{year}{2024}\natexlab{}.
\newblock \showarticletitle{Learning Joint 2-D and 3-D Graph Diffusion Models
  for Complete Molecule Generation}.
\newblock \bibinfo{journal}{\emph{IEEE Transactions on Neural Networks and
  Learning Systems}} \bibinfo{volume}{35}, \bibinfo{number}{9}
  (\bibinfo{year}{2024}), \bibinfo{pages}{11857--11871}.
\newblock


\bibitem[Ji et~al\mbox{.}(2022)]%
        {ji2021survey}
\bibfield{author}{\bibinfo{person}{Shaoxiong Ji}, \bibinfo{person}{Shirui Pan},
  \bibinfo{person}{Erik Cambria}, \bibinfo{person}{Pekka Marttinen}, {and}
  \bibinfo{person}{Philip~S. Yu}.} \bibinfo{year}{2022}\natexlab{}.
\newblock \showarticletitle{A Survey on Knowledge Graphs: Representation,
  Acquisition, and Applications}.
\newblock \bibinfo{journal}{\emph{IEEE Transactions on Neural Networks and
  Learning Systems}} \bibinfo{volume}{33}, \bibinfo{number}{2}
  (\bibinfo{year}{2022}), \bibinfo{pages}{494--514}.
\newblock


\bibitem[Kok and Domingos(2007)]%
        {kok2007statistical}
\bibfield{author}{\bibinfo{person}{Stanley Kok} {and} \bibinfo{person}{Pedro
  Domingos}.} \bibinfo{year}{2007}\natexlab{}.
\newblock \showarticletitle{Statistical Predicate Invention}. In
  \bibinfo{booktitle}{\emph{Proceedings of the 24th International Conference on
  Machine Learning ({ICML})}}. \bibinfo{publisher}{ACM},
  \bibinfo{pages}{433--440}.
\newblock


\bibitem[Kou et~al\mbox{.}(2026)]%
        {kou2026pu}
\bibfield{author}{\bibinfo{person}{Zhiqiang Kou}, \bibinfo{person}{Junyang
  Chen}, \bibinfo{person}{Xin-Qiang Cai}, \bibinfo{person}{Xiaobo Xia},
  \bibinfo{person}{Ming-Kun Xie}, \bibinfo{person}{Dong-Dong Wu},
  \bibinfo{person}{Biao Liu}, \bibinfo{person}{Yuheng Jia},
  \bibinfo{person}{Xin Geng}, \bibinfo{person}{Masashi Sugiyama},
  {et~al\mbox{.}}} \bibinfo{year}{2026}\natexlab{}.
\newblock \showarticletitle{Positive-Unlabeled Reinforcement Learning
  Distillation for On-Premise Small Models}.
\newblock \bibinfo{journal}{\emph{arXiv preprint arXiv:2601.20687}}
  (\bibinfo{year}{2026}).
\newblock


\bibitem[Liu et~al\mbox{.}(2025)]%
        {EvoPath2025}
\bibfield{author}{\bibinfo{person}{Shixuan Liu}, \bibinfo{person}{Haoxiang
  Cheng}, \bibinfo{person}{Yunfei Wang}, \bibinfo{person}{Yue He},
  \bibinfo{person}{Changjun Fan}, {and} \bibinfo{person}{Zhong Liu}.}
  \bibinfo{year}{2025}\natexlab{}.
\newblock \showarticletitle{{EvoPath}: Evolutionary Meta-Path Discovery with
  Large Language Models for Complex Heterogeneous Information Networks}.
\newblock \bibinfo{journal}{\emph{Information Processing \& Management}}
  \bibinfo{volume}{62}, \bibinfo{number}{1} (\bibinfo{year}{2025}),
  \bibinfo{pages}{103920}.
\newblock


\bibitem[Liu et~al\mbox{.}(2024)]%
        {liu2023inductive}
\bibfield{author}{\bibinfo{person}{Shixuan Liu}, \bibinfo{person}{Changjun
  Fan}, \bibinfo{person}{Kewei Cheng}, \bibinfo{person}{Yunfei Wang},
  \bibinfo{person}{Peng Cui}, \bibinfo{person}{Yizhou Sun}, {and}
  \bibinfo{person}{Zhong Liu}.} \bibinfo{year}{2024}\natexlab{}.
\newblock \showarticletitle{Inductive meta-path learning for schema-complex
  heterogeneous information networks}.
\newblock \bibinfo{journal}{\emph{IEEE Transactions on Pattern Analysis and
  Machine Intelligence}} (\bibinfo{year}{2024}).
\newblock


\bibitem[Long et~al\mbox{.}(2024a)]%
        {long2024fact}
\bibfield{author}{\bibinfo{person}{Xiao Long}, \bibinfo{person}{Liansheng
  Zhuang}, \bibinfo{person}{Aodi Li}, \bibinfo{person}{Houqiang Li}, {and}
  \bibinfo{person}{Shafei Wang}.} \bibinfo{year}{2024}\natexlab{a}.
\newblock \showarticletitle{Fact Embedding through Diffusion Model for
  Knowledge Graph Completion}. In \bibinfo{booktitle}{\emph{Proceedings of the
  ACM Web Conference ({WWW})}}. \bibinfo{publisher}{ACM},
  \bibinfo{pages}{2020--2029}.
\newblock


\bibitem[Long et~al\mbox{.}(2024b)]%
        {long2024kgdm}
\bibfield{author}{\bibinfo{person}{Xiao Long}, \bibinfo{person}{Liansheng
  Zhuang}, \bibinfo{person}{Aodi Li}, \bibinfo{person}{Jiuchang Wei},
  \bibinfo{person}{Houqiang Li}, {and} \bibinfo{person}{Shafei Wang}.}
  \bibinfo{year}{2024}\natexlab{b}.
\newblock \showarticletitle{{KGDM}: A Diffusion Model to Capture Multiple
  Relation Semantics for Knowledge Graph Embedding}. In
  \bibinfo{booktitle}{\emph{Proceedings of the AAAI Conference on Artificial
  Intelligence}}, Vol.~\bibinfo{volume}{38}. \bibinfo{pages}{8850--8858}.
\newblock


\bibitem[Luo et~al\mbox{.}(2025)]%
        {luo2023chatrule}
\bibfield{author}{\bibinfo{person}{Linhao Luo}, \bibinfo{person}{Jiaxin Ju},
  \bibinfo{person}{Bo Xiong}, \bibinfo{person}{Yuan-Fang Li},
  \bibinfo{person}{Gholamreza Haffari}, {and} \bibinfo{person}{Shirui Pan}.}
  \bibinfo{year}{2025}\natexlab{}.
\newblock \showarticletitle{{ChatRule}: Mining Logical Rules with Large
  Language Models for Knowledge Graph Reasoning}. In
  \bibinfo{booktitle}{\emph{Proceedings of the Pacific-Asia Conference on
  Knowledge Discovery and Data Mining ({PAKDD})}}.
  \bibinfo{publisher}{Springer}, \bibinfo{pages}{342--355}.
\newblock


\bibitem[Meilicke et~al\mbox{.}(2024)]%
        {AnyBURL2023}
\bibfield{author}{\bibinfo{person}{Christian Meilicke},
  \bibinfo{person}{Melisachew~Wudage Chekol}, \bibinfo{person}{Patrick Betz},
  \bibinfo{person}{Manuel Fink}, {and} \bibinfo{person}{Heiner
  Stuckenschmidt}.} \bibinfo{year}{2024}\natexlab{}.
\newblock \showarticletitle{Anytime bottom-up rule learning for large-scale
  knowledge graph completion: C. Meilicke et al.}
\newblock \bibinfo{journal}{\emph{The VLDB Journal}} \bibinfo{volume}{33},
  \bibinfo{number}{1} (\bibinfo{year}{2024}), \bibinfo{pages}{131--161}.
\newblock


\bibitem[Minervini et~al\mbox{.}(2018)]%
        {minervini2018neuraltheoremprovingscale}
\bibfield{author}{\bibinfo{person}{Pasquale Minervini}, \bibinfo{person}{Matko
  Bo{\v{s}}njak}, \bibinfo{person}{Tim Rockt{\"a}schel}, {and}
  \bibinfo{person}{Sebastian Riedel}.} \bibinfo{year}{2018}\natexlab{}.
\newblock \showarticletitle{Towards Neural Theorem Proving at Scale}. In
  \bibinfo{booktitle}{\emph{Proceedings of the {ICML} Workshop on Neural
  Abstract Machines and Program Induction ({NAMPI})}}.
\newblock


\bibitem[Minervini et~al\mbox{.}(2020)]%
        {minervini2020learningreasoningstrategiesendtoend}
\bibfield{author}{\bibinfo{person}{Pasquale Minervini},
  \bibinfo{person}{Sebastian Riedel}, \bibinfo{person}{Pontus Stenetorp},
  \bibinfo{person}{Edward Grefenstette}, {and} \bibinfo{person}{Tim
  Rockt{\"a}schel}.} \bibinfo{year}{2020}\natexlab{}.
\newblock \showarticletitle{Learning Reasoning Strategies in End-to-End
  Differentiable Proving}. In \bibinfo{booktitle}{\emph{Proceedings of the 37th
  International Conference on Machine Learning ({ICML})}}.
  \bibinfo{publisher}{PMLR}, \bibinfo{pages}{6938--6949}.
\newblock


\bibitem[Pavlovi{\'c} and Sallinger(2024)]%
        {pavlovic2024speede}
\bibfield{author}{\bibinfo{person}{Aleksandar Pavlovi{\'c}} {and}
  \bibinfo{person}{Emanuel Sallinger}.} \bibinfo{year}{2024}\natexlab{}.
\newblock \showarticletitle{{SpeedE}: Euclidean Geometric Knowledge Graph
  Embedding Strikes Back}. In \bibinfo{booktitle}{\emph{Findings of the
  Association for Computational Linguistics: {NAACL} 2024 ({NAACL} Findings)}}.
  \bibinfo{publisher}{Association for Computational Linguistics},
  \bibinfo{pages}{89--98}.
\newblock


\bibitem[Perez et~al\mbox{.}(2018)]%
        {perez2018film}
\bibfield{author}{\bibinfo{person}{Ethan Perez}, \bibinfo{person}{Florian
  Strub}, \bibinfo{person}{Harm de Vries}, \bibinfo{person}{Vincent Dumoulin},
  {and} \bibinfo{person}{Aaron Courville}.} \bibinfo{year}{2018}\natexlab{}.
\newblock \showarticletitle{{FiLM}: Visual Reasoning with a General
  Conditioning Layer}. In \bibinfo{booktitle}{\emph{Proceedings of the AAAI
  Conference on Artificial Intelligence ({AAAI})}}, Vol.~\bibinfo{volume}{32}.
\newblock


\bibitem[Pu et~al\mbox{.}(2024)]%
        {pu2024solving}
\bibfield{author}{\bibinfo{person}{Tianle Pu}, \bibinfo{person}{Chao Chen},
  \bibinfo{person}{Li Zeng}, \bibinfo{person}{Shixuan Liu},
  \bibinfo{person}{Rui Sun}, {and} \bibinfo{person}{Changjun Fan}.}
  \bibinfo{year}{2024}\natexlab{}.
\newblock \showarticletitle{Solving Combinatorial Optimization Problem over
  Graph through {QUBO} Transformation and Deep Reinforcement Learning}. In
  \bibinfo{booktitle}{\emph{Proceedings of the IEEE International Conference on
  Data Mining ({ICDM})}}. \bibinfo{publisher}{IEEE}, \bibinfo{pages}{390--399}.
\newblock


\bibitem[Qu et~al\mbox{.}(2021)]%
        {qu2020rnnlogic}
\bibfield{author}{\bibinfo{person}{Meng Qu}, \bibinfo{person}{Junkun Chen},
  \bibinfo{person}{Louis-Pascal Xhonneux}, \bibinfo{person}{Yoshua Bengio},
  {and} \bibinfo{person}{Jian Tang}.} \bibinfo{year}{2021}\natexlab{}.
\newblock \showarticletitle{{RNNLogic}: Learning Logic Rules for Reasoning on
  Knowledge Graphs}. In \bibinfo{booktitle}{\emph{Proceedings of the
  International Conference on Learning Representations ({ICLR})}}.
\newblock


\bibitem[Quinlan(1990)]%
        {Quinlan1990LearningLD}
\bibfield{author}{\bibinfo{person}{J.~Ross Quinlan}.}
  \bibinfo{year}{1990}\natexlab{}.
\newblock \showarticletitle{Learning Logical Definitions from Relations}.
\newblock \bibinfo{journal}{\emph{Machine Learning}}  \bibinfo{volume}{5}
  (\bibinfo{year}{1990}), \bibinfo{pages}{239--266}.
\newblock


\bibitem[Rockt{\"a}schel and Riedel(2017)]%
        {rocktäschel2017endtoenddifferentiableproving}
\bibfield{author}{\bibinfo{person}{Tim Rockt{\"a}schel} {and}
  \bibinfo{person}{Sebastian Riedel}.} \bibinfo{year}{2017}\natexlab{}.
\newblock \showarticletitle{End-to-End Differentiable Proving}. In
  \bibinfo{booktitle}{\emph{Advances in Neural Information Processing Systems
  ({NeurIPS})}}, Vol.~\bibinfo{volume}{30}.
\newblock


\bibitem[Sadeghian et~al\mbox{.}(2019)]%
        {sadeghian2019drum}
\bibfield{author}{\bibinfo{person}{Ali Sadeghian},
  \bibinfo{person}{Mohammadreza Armandpour}, \bibinfo{person}{Patrick Ding},
  {and} \bibinfo{person}{Daisy~Zhe Wang}.} \bibinfo{year}{2019}\natexlab{}.
\newblock \showarticletitle{{DRUM}: End-to-End Differentiable Rule Mining on
  Knowledge Graphs}. In \bibinfo{booktitle}{\emph{Advances in Neural
  Information Processing Systems ({NeurIPS})}}, Vol.~\bibinfo{volume}{32}.
\newblock


\bibitem[Saxena et~al\mbox{.}(2022)]%
        {saxena-etal-2022-sequence}
\bibfield{author}{\bibinfo{person}{Apoorv Saxena}, \bibinfo{person}{Adrian
  Kochsiek}, {and} \bibinfo{person}{Rainer Gemulla}.}
  \bibinfo{year}{2022}\natexlab{}.
\newblock \showarticletitle{Sequence-to-Sequence Knowledge Graph Completion and
  Question Answering}. In \bibinfo{booktitle}{\emph{Proceedings of the 60th
  Annual Meeting of the Association for Computational Linguistics ({ACL})}}.
  \bibinfo{publisher}{Association for Computational Linguistics},
  \bibinfo{pages}{2814--2828}.
\newblock


\bibitem[Suchanek et~al\mbox{.}(2007)]%
        {suchanek2007yago}
\bibfield{author}{\bibinfo{person}{Fabian~M. Suchanek},
  \bibinfo{person}{Gjergji Kasneci}, {and} \bibinfo{person}{Gerhard Weikum}.}
  \bibinfo{year}{2007}\natexlab{}.
\newblock \showarticletitle{{YAGO}: A Core of Semantic Knowledge}. In
  \bibinfo{booktitle}{\emph{Proceedings of the 16th International Conference on
  World Wide Web ({WWW})}}. \bibinfo{publisher}{ACM},
  \bibinfo{pages}{697--706}.
\newblock


\bibitem[Sun et~al\mbox{.}(2025)]%
        {sun2025kerag}
\bibfield{author}{\bibinfo{person}{Yushi Sun}, \bibinfo{person}{Kai Sun},
  \bibinfo{person}{Yifan~Ethan Xu}, \bibinfo{person}{Xiao Yang},
  \bibinfo{person}{Xin~Luna Dong}, \bibinfo{person}{Nan Tang}, {and}
  \bibinfo{person}{Lei Chen}.} \bibinfo{year}{2025}\natexlab{}.
\newblock \showarticletitle{KERAG: Knowledge-Enhanced Retrieval-Augmented
  Generation for Advanced Question Answering}. In
  \bibinfo{booktitle}{\emph{Findings of the Association for Computational
  Linguistics: EMNLP 2025}}. \bibinfo{pages}{6194--6216}.
\newblock


\bibitem[Sun et~al\mbox{.}(2019)]%
        {sun2019rotate}
\bibfield{author}{\bibinfo{person}{Zhiqing Sun}, \bibinfo{person}{Zhi-Hong
  Deng}, \bibinfo{person}{Jian-Yun Nie}, {and} \bibinfo{person}{Jian Tang}.}
  \bibinfo{year}{2019}\natexlab{}.
\newblock \showarticletitle{{RotatE}: Knowledge Graph Embedding by Relational
  Rotation in Complex Space}. In \bibinfo{booktitle}{\emph{Proceedings of the
  International Conference on Learning Representations ({ICLR})}}.
\newblock


\bibitem[Toutanova and Chen(2015)]%
        {toutanova2015observed}
\bibfield{author}{\bibinfo{person}{Kristina Toutanova} {and}
  \bibinfo{person}{Danqi Chen}.} \bibinfo{year}{2015}\natexlab{}.
\newblock \showarticletitle{Observed versus Latent Features for Knowledge Base
  and Text Inference}. In \bibinfo{booktitle}{\emph{Proceedings of the 3rd
  Workshop on Continuous Vector Space Models and their Compositionality
  ({CVSC})}}. \bibinfo{publisher}{Association for Computational Linguistics},
  \bibinfo{pages}{57--66}.
\newblock


\bibitem[Trouillon et~al\mbox{.}(2016)]%
        {trouillon2016complex}
\bibfield{author}{\bibinfo{person}{Th{\'e}o Trouillon},
  \bibinfo{person}{Johannes Welbl}, \bibinfo{person}{Sebastian Riedel},
  \bibinfo{person}{{\'E}ric Gaussier}, {and} \bibinfo{person}{Guillaume
  Bouchard}.} \bibinfo{year}{2016}\natexlab{}.
\newblock \showarticletitle{Complex Embeddings for Simple Link Prediction}. In
  \bibinfo{booktitle}{\emph{Proceedings of the 33rd International Conference on
  Machine Learning ({ICML})}}. \bibinfo{publisher}{PMLR},
  \bibinfo{pages}{2071--2080}.
\newblock


\bibitem[Vashishth et~al\mbox{.}(2020)]%
        {vashishth2019composition}
\bibfield{author}{\bibinfo{person}{Shikhar Vashishth}, \bibinfo{person}{Soumya
  Sanyal}, \bibinfo{person}{Vikram Nitin}, {and} \bibinfo{person}{Partha
  Talukdar}.} \bibinfo{year}{2020}\natexlab{}.
\newblock \showarticletitle{Composition-Based Multi-Relational Graph
  Convolutional Networks}. In \bibinfo{booktitle}{\emph{Proceedings of the
  International Conference on Learning Representations ({ICLR})}}.
\newblock


\bibitem[Vignac et~al\mbox{.}(2023)]%
        {vignac2022digress}
\bibfield{author}{\bibinfo{person}{Cl{\'e}ment Vignac}, \bibinfo{person}{Igor
  Krawczuk}, \bibinfo{person}{Antoine Siraudin}, \bibinfo{person}{Bohan Wang},
  \bibinfo{person}{Volkan Cevher}, {and} \bibinfo{person}{Pascal Frossard}.}
  \bibinfo{year}{2023}\natexlab{}.
\newblock \showarticletitle{{DiGress}: Discrete Denoising Diffusion for Graph
  Generation}. In \bibinfo{booktitle}{\emph{Proceedings of the International
  Conference on Learning Representations ({ICLR})}}.
\newblock


\bibitem[Watson et~al\mbox{.}(2023)]%
        {watson2023novo}
\bibfield{author}{\bibinfo{person}{Joseph~L Watson}, \bibinfo{person}{David
  Juergens}, \bibinfo{person}{Nathaniel~R Bennett}, \bibinfo{person}{Brian~L
  Trippe}, \bibinfo{person}{Jason Yim}, \bibinfo{person}{Helen~E Eisenach},
  \bibinfo{person}{Woody Ahern}, \bibinfo{person}{Andrew~J Borst},
  \bibinfo{person}{Robert~J Ragotte}, \bibinfo{person}{Lukas~F Milles},
  {et~al\mbox{.}}} \bibinfo{year}{2023}\natexlab{}.
\newblock \showarticletitle{De novo design of protein structure and function
  with RFdiffusion}.
\newblock \bibinfo{journal}{\emph{Nature}} \bibinfo{volume}{620},
  \bibinfo{number}{7976} (\bibinfo{year}{2023}), \bibinfo{pages}{1089--1100}.
\newblock


\bibitem[Yang et~al\mbox{.}(2015)]%
        {yang2014embedding}
\bibfield{author}{\bibinfo{person}{Bishan Yang}, \bibinfo{person}{Wen-tau Yih},
  \bibinfo{person}{Xiaodong He}, \bibinfo{person}{Jianfeng Gao}, {and}
  \bibinfo{person}{Li Deng}.} \bibinfo{year}{2015}\natexlab{}.
\newblock \showarticletitle{Embedding Entities and Relations for Learning and
  Inference in Knowledge Bases}. In \bibinfo{booktitle}{\emph{Proceedings of
  the International Conference on Learning Representations ({ICLR})}}.
\newblock


\bibitem[Yang et~al\mbox{.}(2017)]%
        {yang2017differentiable}
\bibfield{author}{\bibinfo{person}{Fan Yang}, \bibinfo{person}{Zhilin Yang},
  {and} \bibinfo{person}{William~W. Cohen}.} \bibinfo{year}{2017}\natexlab{}.
\newblock \showarticletitle{Differentiable Learning of Logical Rules for
  Knowledge Base Reasoning}. In \bibinfo{booktitle}{\emph{Advances in Neural
  Information Processing Systems ({NeurIPS})}}, Vol.~\bibinfo{volume}{30}.
  \bibinfo{pages}{2316--2325}.
\newblock


\bibitem[Yang et~al\mbox{.}(2026)]%
        {yang2026mad}
\bibfield{author}{\bibinfo{person}{Haocheng Yang}, \bibinfo{person}{Fengxiang
  Cheng}, \bibinfo{person}{Tianjun Yao}, \bibinfo{person}{Mengyue Yang},
  \bibinfo{person}{Jiajun Chai}, \bibinfo{person}{Xiaohan Wang},
  \bibinfo{person}{Guojun Yin}, \bibinfo{person}{Wei Lin},
  \bibinfo{person}{Soummya Kar}, \bibinfo{person}{Fenrong Liu},
  \bibinfo{person}{Haoxuan Li}, {and} \bibinfo{person}{Yisen Wang}.}
  \bibinfo{year}{2026}\natexlab{}.
\newblock \showarticletitle{{MAD}-{L}ogic: Multi-Agent Debate Enhances Symbolic
  Translation and Reasoning}. In \bibinfo{booktitle}{\emph{The Fourteenth
  International Conference on Learning Representations}}.
\newblock


\bibitem[Yang et~al\mbox{.}(2023)]%
        {yang2023diffusion}
\bibfield{author}{\bibinfo{person}{Ling Yang}, \bibinfo{person}{Zhilong Zhang},
  \bibinfo{person}{Yang Song}, \bibinfo{person}{Shenda Hong},
  \bibinfo{person}{Runsheng Xu}, \bibinfo{person}{Yue Zhao},
  \bibinfo{person}{Wentao Zhang}, \bibinfo{person}{Bin Cui}, {and}
  \bibinfo{person}{Ming-Hsuan Yang}.} \bibinfo{year}{2023}\natexlab{}.
\newblock \showarticletitle{Diffusion Models: A Comprehensive Survey of Methods
  and Applications}.
\newblock \bibinfo{journal}{\emph{Comput. Surveys}} \bibinfo{volume}{56},
  \bibinfo{number}{4} (\bibinfo{year}{2023}), \bibinfo{pages}{1--39}.
\newblock


\bibitem[Yin et~al\mbox{.}(2025)]%
        {yin2025efokcqa}
\bibfield{author}{\bibinfo{person}{Hang Yin}, \bibinfo{person}{Zihao Wang},
  \bibinfo{person}{Weizhi Fei}, {and} \bibinfo{person}{Yangqiu Song}.}
  \bibinfo{year}{2025}\natexlab{}.
\newblock \showarticletitle{{EFO}$_k$-{CQA}: Towards Knowledge Graph Complex
  Query Answering beyond Set Operation}. In
  \bibinfo{booktitle}{\emph{Proceedings of the 31st ACM SIGKDD Conference on
  Knowledge Discovery and Data Mining}}. \bibinfo{publisher}{ACM},
  \bibinfo{pages}{5876--5887}.
\newblock


\bibitem[Zhang et~al\mbox{.}(2024)]%
        {zhang2024question}
\bibfield{author}{\bibinfo{person}{Yu Zhang}, \bibinfo{person}{Kehai Chen},
  \bibinfo{person}{Xuefeng Bai}, \bibinfo{person}{Zhao Kang},
  \bibinfo{person}{Quanjiang Guo}, {and} \bibinfo{person}{Min Zhang}.}
  \bibinfo{year}{2024}\natexlab{}.
\newblock \showarticletitle{Question-guided knowledge graph re-scoring and
  injection for knowledge graph question answering}. In
  \bibinfo{booktitle}{\emph{Findings of the Association for Computational
  Linguistics: EMNLP 2024}}. \bibinfo{pages}{8972--8985}.
\newblock


\bibitem[Zhou et~al\mbox{.}(2024)]%
        {zhou2024unifying}
\bibfield{author}{\bibinfo{person}{Cai Zhou}, \bibinfo{person}{Xiyuan Wang},
  {and} \bibinfo{person}{Muhan Zhang}.} \bibinfo{year}{2024}\natexlab{}.
\newblock \showarticletitle{Unifying Generation and Prediction on Graphs with
  Latent Graph Diffusion}. In \bibinfo{booktitle}{\emph{Advances in Neural
  Information Processing Systems ({NeurIPS})}}, Vol.~\bibinfo{volume}{37}.
  \bibinfo{pages}{61963--61999}.
\newblock


\bibitem[Zhu et~al\mbox{.}(2023)]%
        {zhu2023net}
\bibfield{author}{\bibinfo{person}{Zhaocheng Zhu}, \bibinfo{person}{Xinyu
  Yuan}, \bibinfo{person}{Michael Galkin}, \bibinfo{person}{Louis-Pascal
  Xhonneux}, \bibinfo{person}{Ming Zhang}, \bibinfo{person}{Maxime Gazeau},
  {and} \bibinfo{person}{Jian Tang}.} \bibinfo{year}{2023}\natexlab{}.
\newblock \showarticletitle{{A}*Net: A Scalable Path-Based Reasoning Approach
  for Knowledge Graphs}. In \bibinfo{booktitle}{\emph{Advances in Neural
  Information Processing Systems ({NeurIPS})}}, Vol.~\bibinfo{volume}{36}.
  \bibinfo{pages}{59323--59336}.
\newblock


\bibitem[Zhu et~al\mbox{.}(2021)]%
        {zhu2021neural}
\bibfield{author}{\bibinfo{person}{Zhaocheng Zhu}, \bibinfo{person}{Zuobai
  Zhang}, \bibinfo{person}{Louis-Pascal Xhonneux}, {and} \bibinfo{person}{Jian
  Tang}.} \bibinfo{year}{2021}\natexlab{}.
\newblock \showarticletitle{Neural Bellman-Ford Networks: A General Graph
  Neural Network Framework for Link Prediction}. In
  \bibinfo{booktitle}{\emph{Advances in Neural Information Processing Systems
  ({NeurIPS})}}, Vol.~\bibinfo{volume}{34}. \bibinfo{pages}{29476--29490}.
\newblock


\end{thebibliography}

\appendix
\renewcommand{\thesection}{Appendix~\Alph{section}}
\renewcommand{\thesubsection}{\Alph{section}.\arabic{subsection}}
\renewcommand{\thesubsubsection}{\Alph{section}.\arabic{subsection}.\arabic{subsubsection}}

\section*{Appendices}
\addcontentsline{toc}{section}{Appendix}

\section{Dataset Description}
\label{sec:dataset}

\begin{itemize}
\item \textbf{Family}~\cite{hinton1986learning} captures relations among family members.
\item \textbf{Kinship}~\cite{kok2007statistical} contains kinship relations among members of the Alyawarra tribe from
Central Australia.
\item \textbf{UMLS}~\cite{kok2007statistical} contains biomedical concepts as entities and treatment and diagnosis relations.
\item \textbf{WN-18RR}~\cite{dettmers2018convolutional} is a lexical dataset whose entities refer to word senses and whose relations define lexical connections.
\item \textbf{FB15K-237}~\cite{toutanova2015observed} is derived from Freebase, an online collection of structured data extracted from multiple sources.

\item \textbf{YAGO3-10}~\cite{suchanek2007yago} is a subset of YAGO, a large semantic knowledge base constructed from multiple sources.
\end{itemize}

\begin{table*}[t]

\centering
\caption{Representative chain-like and graph-like rules learned by \model~on YAGO3-10.}
\begin{tabular}{ccl}
\toprule[1pt]
\textbf{Rule Type} & \textbf{Rule Head} & \textbf{Rule Body} \\
\midrule
\multirow{6}{*}{\textbf{Chain-like}}
& \multirow{3}{*}{isLocatedIn}
& hasNeighbor$(x,z) \wedge $isLocatedIn$(z,y)$ \\
&
& hasNeighbor$(x,z_1) \wedge $DealsWith$(z_1,z_2) \wedge $isLocatedIn$(z_2,y)$ \\
&
& hasCapital$(x,z) \wedge $participatedIn$(x,z) \wedge $happenedIn$(z,y)$ \\
\cmidrule(lr){2-3}
& \multirow{3}{*}{isCitizenOf}
& LivesIn$(x,y)$ \\
&
& isMarriedTo$(x,z) \wedge $isCitizenOf$(z,y)$ \\
&
& hasAcademicAdvisor$(x,z_1) \wedge $worksAt$(z_1,z_2) \wedge $isLocatedIn$(z_2,y)$ \\
\midrule
\multirow{6}{*}{\textbf{Graph-like}}
& \multirow{3}{*}{isLocatedIn}
& hasCapital$(x,z_1) \wedge $hasCapital$(y,z_1) \wedge $hasOfficialLanguage$(x,z_2) \wedge $hasOfficialLanguage$(y,z_2)$ \\
&
& exports$(x,z_1) \wedge $imports$(y,z_1) \wedge $hasCurrency$(x,z_2) \wedge $hasCurrency$(y,z_2)$ \\
&
& hasNeighbor$(x,z_1) \wedge $hasNeighbor$(y,z_1) \wedge $exports$(x,z_2) \wedge $imports$(y,z_2)$ \\
\cmidrule(lr){2-3}
& \multirow{3}{*}{isCitizenOf}
& LivesIn$(x,y) \wedge $WorksAt$(x,y)$ \\
&
& isMarriedTo$(x,z_1) \wedge $isCitizenOf$(z_1,z_2) \wedge $DealsWith$(z_1,y)$ \\
&
& GraduatedFrom$(x,z_1) \wedge $isLocatedIn$(z_1,y) \wedge $hasAcademicAdvisor$(x,z_2) \wedge $livesIn$(z_2,y)$ \\
\bottomrule[1pt]
\end{tabular}
\label{table_rule-examples}
\end{table*}

\begin{table}[t]
\centering

\caption{Performance evaluation results under different random seeds across various datasets.}
\label{tab:seed_comparison}
\resizebox{\linewidth}{!}{
\begin{tabular}{lccccccccc}
\toprule[1pt]
 \multirow{2}{*}{Seed} & \multicolumn{3}{c}{Family} & \multicolumn{3}{c}{Kinship} & \multicolumn{3}{c}{UMLS} \\
\cmidrule(lr){2-4} \cmidrule(lr){5-7} \cmidrule(lr){8-10}
 & MRR (\%) & HIT1 (\%) & HIT10 (\%) & MRR (\%) & HIT1 (\%) & HIT10 (\%) & MRR (\%) & HIT1 (\%) & HIT10 (\%) \\
\midrule
42   & 95.38 & 91.75 & 99.64 & 68.33 & 56.46 & 92.43 & 83.45 & 74.96 & 95.60 \\
123  & 94.97 & 91.11 & 99.61 & 67.02 & 53.90 & 92.67 & 84.01 & 75.72 & 97.12 \\
456  & 95.31 & 91.60 & 99.71 & 66.26 & 52.27 & 92.67 & 83.57 & 74.66 & 96.81 \\
789  & 95.20 & 91.38 & 93.00 & 67.14 & 54.48 & 92.43 & 82.93 & 73.97 & 97.11 \\
1024 & 95.39 & 91.86 & 99.64 & 67.17 & 54.71 & 92.08 & 83.43 & 75.11 & 95.75 \\
\midrule
MEAN & 95.25 & 91.54 & 98.32 & 67.18 & 54.36 & 92.46 & 83.48 & 74.89 & 96.48 \\
95\% CI & $\pm 0.22$ & $\pm 0.37$ & $\pm 3.69$ & $\pm 0.92$ & $\pm 1.88$ & $\pm 0.30$ & $\pm 0.48$ & $\pm 0.80$ & $\pm 0.93$ \\
VAR  & 0.00  & 0.00  & 0.09  & 0.00  & 0.02  & 0.00  & 0.35  & 0.57  & 0.67  \\
\bottomrule[1pt]
\end{tabular}}

\vspace{0.5em} 

\resizebox{\linewidth}{!}{
\begin{tabular}{lccccccccc}
\toprule[1pt]
\multirow{2}{*}{Seed} & \multicolumn{3}{c}{WN-18RR} & \multicolumn{3}{c}{FB15K-237} & \multicolumn{3}{c}{YAGO3-10} \\
\cmidrule(lr){2-4} \cmidrule(lr){5-7} \cmidrule(lr){8-10}
 & MRR (\%) & HIT1 (\%) & HIT10 (\%) & MRR (\%) & HIT1 (\%) & HIT10 (\%) & MRR (\%) & HIT1 (\%) & HIT10 (\%) \\
\midrule
42   & 45.27 & 40.26 & 54.66 & 42.31 & 32.32 & 61.72 & 55.70 & 50.42 & 63.80 \\
123  & 43.91 & 39.40 & 52.50 & 42.61 & 32.16 & 63.04 & 56.78 & 51.64 & 65.23 \\
456  & 44.24 & 39.80 & 52.73 & 42.41 & 32.04 & 62.51 & 52.53 & 47.31 & 60.66 \\
789  & 42.73 & 38.30 & 51.09 & 42.63 & 32.33 & 62.18 & 54.93 & 49.86 & 62.71 \\
1024 & 42.74 & 38.33 & 50.85 & 41.49 & 31.40 & 61.17 & 54.62 & 49.65 & 62.47 \\
\midrule
MEAN & 43.78 & 39.22 & 52.37 & 42.29 & 32.05 & 62.13 & 54.91 & 49.78 & 62.97 \\
95\% CI & $\pm 1.34$ & $\pm 1.09$ & $\pm 1.90$ & $\pm 0.58$ & $\pm 0.48$ & $\pm 0.89$ & $\pm 1.95$ & $\pm 1.96$ & $\pm 2.10$ \\
VAR  & 0.01  & 0.01  & 0.02  & 0.42  & 0.34  & 0.64  & 1.41  & 1.41  & 1.51  \\
\bottomrule[1pt]
\end{tabular}
}
\end{table}
\section{Baseline Description}
\label{sec:baseline_appendix}

\textbf{Embedding Baselines.}

    \begin{itemize}
    \item \textbf{TransE}~\cite{bordes2013transe}.   TransE models relations as translations between entity embeddings, capturing  one-hop relational patterns.

    \item \textbf{DistMult}~\cite{yang2014embedding}. DistMult uses diagonal bilinear scoring for entity-relation interactions.

    \item \textbf{ComplEx}~\cite{trouillon2016complex}. ComplEx uses complex embeddings to capture symmetric and antisymmetric relations.

    \item \textbf{RotatE}~\cite{sun2019rotate}. RotatE models relations as rotations in complex space and captures inversion patterns.

    \item \textbf{SpeedE}~\cite{pavlovic2024speede}. SpeedE is a lightweight Euclidean geometric KGE model for efficient KGC.
    \end{itemize}

\textbf{GNN-based Baselines.}
    \begin{itemize}

        \item \textbf{CompGCN}~\citep{vashishth2019composition}. CompGCN jointly embeds nodes and relations via compositional GCN aggregation.
        
        \item \textbf{NBFNet}~\cite{zhu2021neural}. NBFNet learns Bellman-Ford-style path representations for link prediction through neural path aggregation over relation-specific message passing.
        
        \item \textbf{A$^{*}$Net}~\cite{zhu2023net}. A$^{*}$Net extends NBFNet with A$^{*}$-like heuristic search to prioritize informative paths and reduce unnecessary expansion during reasoning over large KGs.
    \end{itemize}

\textbf{Rule-based Baselines.}
    \begin{itemize}
         \item \textbf{AnyBURL}~\cite{AnyBURL2023}.   AnyBURL mines rules via anytime bottom-up path sampling and confidence-based rule selection.

        \item \textbf{AMIE}~\cite{AMIE2013luis}. AMIE mines Horn rules in large KBs under Partical Compeletness Assumption for scalable rule evaluation.      
    
        \item \textbf{Neural-LP}~\cite{yang2017differentiable}. Neural-LP learns logical rules through differentiable tensor operations.

        \item \textbf{DRUM}~\cite{sadeghian2019drum}. DRUM extends Neural-LP with RNNs to model longer rule chains and recurrent dependencies.

        \item \textbf{RNNLogic}~\cite{qu2020rnnlogic}. RNNLogic uses an RNN-based generator to propose logical rules for KG reasoning.
        
        \item \textbf{RLogic}~\citep{Rlogic2022}. RLogic introduces recursive logical rule learning through predicate representation learning and recursive rule composition.

        \item \textbf{NCRL}~\citep{cheng2023NCRL}.  NCRL is a neural compositional rule learning method that uses attention mechanisms for recursive reasoning over rule components.
       
        \item \textbf{ChatRule}~\cite{luo2023chatrule}. ChatRule uses prompts to guide LLMs in generating logical rules with KG structural validation.

    \end{itemize}





\begin{table}[t]
\centering
\caption{Statistics and properties of the experimental datasets.}
\label{tab:dataset_structure_analysis}
\resizebox{\linewidth}{!}{
\begin{tabular}{lccccccc}
\toprule[1pt]
Dataset & \# Rel & BF\_Schema & Gini & NPIM & N-to-N ratio & Chain Better & Graph Better \\
\midrule
Family    & 12  & 7.02  & 0.20 & 0.55  & 66.70\% & 100.00\% & 0.00\%   \\
Kinship   & 25  & 30.22 & 0.39 & 0.36  & 88.00\% & 72.00\%  & 0.00\%   \\
UMLS      & 46  & 11.48 & 0.64 & 0.36  & 67.40\% & 30.43\%  & 17.39\%  \\
WN-18RR   & 11  & 2.34  & 0.67 & -0.07 & 18.20\% & 63.64\%  & 27.27\%  \\
FB15K-237 & 237 & 10.33 & 0.68 & 0.47  & 45.60\% & 68.78\%  & 20.68\%  \\
YAGO3-10  & 37  & 2.88  & 0.83 & 0.26  & 45.90\% & 70.27\%  & 27.03\%  \\
\bottomrule[1pt]
\end{tabular}
}
\end{table}

\section{Impact of KG Structural Properties}
\label{sec:KG_structure_analysis}
To examine when graph-like rules are useful, we analyze the structural indicators, including schema-level connectivity (BF\_Schema), relation frequency imbalance (Gini), relation co-occurrence tendency (NPIM), and the many-to-many (N-to-N) ratio, in Table~\ref{tab:dataset_structure_analysis} and report Graph Better, the proportion of relations where graph-like rules outperform chain-like rules.

Graph-like rules provide limited gains on Family, Kinship, and WN-18RR for different reasons. Family and Kinship contain semantically well-defined kinship relations, where chain-like rules already capture the main reasoning patterns. WN-18RR is instead limited by weak relation co-occurrence, with many relations being mutually exclusive rather than jointly informative (NPIM = -0.07). Therefore, graph constraints provide limited additional evidence on these datasets.

In contrast, UMLS, FB15K-237, and YAGO3-10 show higher Graph Better values. These datasets have more skewed relation distributions and richer relational ambiguity, where chain-like rules may match multiple candidates. Graph-like rules are therefore more useful because their conjunctive constraints help reduce ambiguity.

Overall, graph-like rules are most beneficial when additional structural constraints can disambiguate chain-based reasoning, but their advantage is limited when relation semantics are already regular or weakly co occurring.
\section{Rule Comparison}
\label{appendix:rule comparision}
Representative chain-like and graph-like rules generated by~\model~on YAGO3-10 are presented in Table~\ref{table_rule-examples}. Here, we analyze graph-like rules according to the four rule-body types in Fig.~\ref{fig3:metagraph_type}.

\noindent \textbf{Linear Rule Body (Fig.~\ref{fig:pic3a}).} This type represents a sequential variable chain. The rule \textit{isCitizenOf}$(x, y) \leftarrow$ \textit{hasAcademicAdvisor}(x, $z_1$) $\wedge$ \textit{worksAt}$(z_1, z_2) \wedge$ \textit{isLocatedIn}$(z_2, y)$ demonstrates this structure, where inference follows a single path from $x$ to $y$ through the workplace of the advisor and its location. This rule captures citizenship through the institutional affiliation of the advisor.

\noindent \textbf{Branching Rule Body (Fig.~\ref{fig:pic3b}).} This type contains a branching node connected to multiple nodes. The rule \textit{isCitizenOf}$(x, y) \leftarrow$ \textit{isMarriedTo}$(x, z_1) \wedge$ \textit{isCitizenOf}$(z_1, z_2) \wedge$ \textit{DealsWith}$(z_1, y)$ branches at node $z_1$, which connects to both $z_2$ and $y$. This rule captures citizenship through familial and relational dependencies.

\noindent \textbf{Edge-Cyclic Rule Body (Fig.~\ref{fig:pic3c}).} This type contains multiple edges between the same entity pair. The rule \textit{isCitizenOf}$(x, y) \leftarrow$ \textit{LivesIn}$(x, y) \wedge$ \textit{WorksAt}$(x, y)$ forms an edge cycle in which two relations connect $x$ and $y$, showing how multiple co-location signals support citizenship inference.

\noindent \textbf{Node-Cyclic Rule Body (Fig.~\ref{fig:pic3d}).} This type contains shared intermediate nodes that form cycles. The rule \textit{isLocatedIn}$(x, y) \leftarrow$ \textit{hasCapital}$(x, z_1) \wedge$\textit{hasCapital}$(y, z_1) \wedge$ \textit{hasOfficialLanguage}$(x, z_2) \wedge$ \textit{hasOfficialLanguage}$(y, z_2)$ connects $x$ and $y$ through the shared capital $z_1$ and language $z_2$, reflecting how shared attributes can indicate hierarchical relationships.

\section{Multi-seed Robustness Analysis}
\label{sec:seed_analysis}

We evaluate~\model~under five random seeds to assess robustness. As shown in Table~\ref{tab:seed_comparison}, the results remain stable across datasets. The 95\% confidence intervals are small, and the variance stays close to zero on small datasets and below 1.5 on more complex datasets. These results indicate that~\model~is not sensitive to random initialization.

\section{KGC Pipeline}
\label{sec:eval_process}

We apply generated rules to KGC through sparse matrix grounding and filtered ranking, illustrated in Algorithm~\ref{alg:kgc_pipeline}. For each relation $r\in\mathcal{R}$, let $\mathbf{M}_r\in\{0,1\}^{|\mathcal{E}|\times|\mathcal{E}|}$ denote its sparse adjacency matrix, and let $\mathbf{M}^{\mathrm{gt}}_r$ denote the ground-truth matrix used for scoring and filtering. The generated rules for $r$ are partitioned into $\Omega_r^{\mathrm{chain}}$ and $\Omega_r^{\mathrm{graph}}$. For a rule $\rho$, $\mathrm{Ground}(\rho)$ returns its body grounding matrix: chain-like rules are grounded by sparse matrix multiplication, while graph-like rules are grounded by decomposing the body into chain components and taking their element-wise conjunction. $\mathrm{Score}(\cdot)$ computes rule quality from coverage, confidence, and PCA-confidence. For each query $(e_h,r,?)$, candidate tail entities are ranked by $\mathbf{S}_r[e_h,:]$ under the filtered setting. We report Hit@K and MRR: $\mathrm{Hit@K}=\frac{1}{|Q|}\sum_{q\in Q}\mathbb{I}(\mathrm{rank}_q\le K), \mathrm{MRR}=\frac{1}{|Q|}\sum_{q\in Q}\frac{1}{\mathrm{rank}_q}.$

\section{Rule Dataset Construction}
\label{appendix:data_construction}

For supervised pre-training, we construct rule instances from KGs in two steps. First, for each training triple $(e_h,r_h,e_t)$, we sample fixed-length random walks from $e_h$ and convert the resulting relation paths into chain-like rule bodies by replacing entities with logical variables. Second, each chain is augmented once into a graph-like rule by sampling one topology from Fig.~\ref{fig3:metagraph_type} and adding the required auxiliary variables and relation labels. Each constructed instance is represented by a rule-body adjacency matrix $\mathbf{A}_0$, with the target relation $r_h$ used as the generation condition.

\begin{algorithm}[t]
\small
\renewcommand\arraystretch{1.15}
\caption{KGC with Generated Rules}
\label{alg:kgc_pipeline}
\begin{algorithmic}[1]
\REQUIRE Matrices $\{\mathbf{M}_r,\mathbf{M}^{\mathrm{gt}}_r\}_{r\in\mathcal{R}}$, rules $\{\Omega_r^{\mathrm{chain}},\Omega_r^{\mathrm{graph}}\}_{r\in\mathcal{R}}$, fusion weight $\alpha$, queries $Q$
\ENSURE Filtered ranks $\{\mathrm{rank}_q\}_{q\in Q}$
\FOR{$r \in \mathcal{R}$}
    \STATE $\mathbf{S}^{\mathrm{chain}}_r,\mathbf{S}^{\mathrm{graph}}_r \leftarrow \mathbf{0},\mathbf{0}$
    \FOR{$\rho \in \Omega_r^{\mathrm{chain}} \cup \Omega_r^{\mathrm{graph}}$}
        \STATE $\mathbf{M}_{\rho} \leftarrow \mathrm{Ground}(\rho)$
        \STATE $s(\rho) \leftarrow \mathrm{Score}(\mathbf{M}_{\rho},\mathbf{M}^{\mathrm{gt}}_r)$
        \STATE $\mathbf{S}^{\mathrm{type}(\rho)}_r \leftarrow \mathbf{S}^{\mathrm{type}(\rho)}_r+s(\rho)\mathbf{M}_{\rho}$
    \ENDFOR
    \STATE $\mathbf{S}_r \leftarrow (1-\alpha)\mathbf{S}^{\mathrm{chain}}_r+\alpha\mathbf{S}^{\mathrm{graph}}_r$
\ENDFOR
\FOR{$q=(e_h,r,e_t^\star)\in Q$}
    \STATE $\mathbf{s}_q \leftarrow \mathbf{S}_r[e_h,:]$
    \STATE Mask all known valid tails except $e_t^\star$ in $\mathbf{s}_q$
    \STATE $\mathrm{rank}_q \leftarrow 1+\sum_{e\in\mathcal{E}}\mathbb{I}\!\left(\mathbf{s}_q[e]>\mathbf{s}_q[e_t^\star]\right)$
\ENDFOR
\RETURN $\{\mathrm{rank}_q\}_{q\in Q}$

\end{algorithmic}
\end{algorithm}

\end{document}